\documentclass[a4paper,fleqn,preprint]{cas-sc}

\usepackage[numbers]{natbib}
\usepackage{amsmath,bm}
\usepackage{caption}
\usepackage{subcaption}
\usepackage{lscape}
\usepackage{comment}
\usepackage{lineno}
\usepackage{graphicx}
\usepackage[flushleft]{threeparttable}
\usepackage{xcolor}
\newcolumntype{P}[1]{>{\centering\arraybackslash}p{#1}}

\definecolor{Blue2}{rgb}{0.3010, 0.7450, 0.9330}
\definecolor{Green}{rgb}{0.4660 0.6740 0.1880}
\definecolor{Lilac}{rgb}{0.4940, 0.1840, 0.5560}
\definecolor{Blue1}{rgb}{0, 0, 1}
\definecolor{Blue3}{rgb}{0.5020, 0.7569, 0.8588}
\definecolor{Red}{rgb}{1, 0, 0}
\definecolor{Black}{rgb}{0, 0, 0}

\def\tsc#1{\csdef{#1}{\textsc{\lowercase{#1}}\xspace}}
\tsc{WGM}
\tsc{QE}
\tsc{EP}
\tsc{PMS}
\tsc{BEC}
\tsc{DE}
\begin{document}
\let\WriteBookmarks\relax
\def\floatpagepagefraction{1}
\def\textpagefraction{.001}
\shorttitle{A Robust Learning Methodology for Uncertainty-aware Scientific Machine Learning models}
\shortauthors{Costa E. A. et~al.}

\title{A robust learning methodology for uncertainty-aware scientific machine learning models}                      
%\tnotemark[1,2]

\tnotetext[1]{This study was financed in part by the CNPq, CAPES (financial code 001), FAPESB and Petrobras.}

%\tnotetext[2]{The second title footnote which is a longer text matter
   %to fill through the whole text width and overflow into
  % another line in the footnotes area of the first page.}

\author[1]{Erbet Almeida Costa}[type=editor,
                        auid=000,bioid=1,
                        role=Researcher,
                        orcid=0000-0003-1397-9628]
%\cormark[1]
\fnmark[1]
\ead{erbetcosta@ufba.br}
\ead[url]{erbetcosta@ufba.br}

\credit{Conceptualization of this study, Methodology, Software, Original draft preparation}

\address[1]{Programa de pós-graduação em Mecatrônica, Universidade Federal da Bahia, Rua Prof. Aristides Novis, 2, Federação, 40210-630, Salvador, Bahia, Brasil.}

\author[2,3]{Carine de Menezes Rebello}[role=Researcher]
\credit{Data curation, Original draft preparation}
\ead[url]{carine.menezes@ufba.br}

\author[1]{Márcio Fontana}[%
   role=Professor
   ]
\fnmark[2]
\ead{mfontana@ufba.br}

\credit{Writing - Original draft preparation}

\address[2]{LSRE-LCM - Laboratory of Separation and Reaction Engineering – Laboratory of Catalysis and Materials, Faculty of Engineering, University of Porto, Rua Dr. Roberto Frias, 4200-465 Porto, Portugal.}

\author[1,3]{Leizer Schnitman}[%
   role=Professor
   ]
%\cormark[2]
\fnmark[1]
\credit{Conceptualization, Methodology, Original draft preparation}

\ead{leizer@ufba.br}

\author[2,3]{Idelfonso Bessa dos Reis Nogueira}[ role=Professor
   ]
%\cormark[3]
\ead{idelfonso@fe.up.pt}
\credit{Conceptualization, Methodology, Original draft preparation}

\address[3]{ALiCE - Associate Laboratory in Chemical Engineering, Faculty of Engineering, University of Porto, Rua Dr. Roberto Frias, 4200-465 Porto, Portugal.}

\cortext[cor1]{First Corresponding author}
\cortext[cor2]{Second corresponding author}
\cortext[cor3]{Third corresponding author}
%fntext[fn1]{This is the first author footnote. but is common to third
%  author as well.}
%\fntext[fn2]{Another author footnote, this is a very long footnote and
 % it should be a really long footnote. But this footnote is not yet
%  sufficiently long enough to make two lines of footnote text.}

%\nonumnote{This note has no numbers. In this work we demonstrate $a_b$
%  the formation Y\_1 of a new type of polariton on the interface
 % between a cuprous oxide slab and a polystyrene micro-sphere placed
 % on the slab.
 % }

\begin{abstract}
Robust learning is an important issue in Scientific Machine Learning (SciML). There are several works in the literature addressing this topic. However, there is an increasing demand for methods that can simultaneously consider all the different uncertainty components involved in SciML model identification. Hence, this work proposes a comprehensive methodology for uncertainty evaluation of the SciML that also considers several possible sources of uncertainties involved in the identification process. The uncertainties considered in the proposed method are the absence of theory and causal models, the sensitiveness to data corruption or imperfection, and the computational effort. Therefore, it was possible to provide an overall strategy for the uncertainty-aware models in the SciML field. The methodology is validated through a case study, developing a Soft Sensor for a polymerization reactor. The results demonstrated that the identified Soft Sensor are robust for uncertainties, corroborating with the consistency of the proposed approach.
\end{abstract}

%\includegraphics{figs/grabs.pdf}

%\item Research highlights item 1
%\item Research highlights item 2
%\item Research highlights item 3

\begin{keywords}
Scientific Machine Learning \sep Robust Learning \sep Uncertainty \sep Markov Chain Monte Carlo
\end{keywords}

\maketitle

\section{Introduction}

Machine learning (ML) has been widespread in several application domains. Hence, giving birth to a new field of study, Scientific Machine Learning (SciML). This field is concerned with the proper application of an ML model, considering all peculiarities of a given domain. As ML becomes more popular, several concerns have risen \citep{Rackauckas,Chuang2018,Gaikwad2020,Nogueira2022}. One issue that can be highlighted is the development of robust machine learning, the so-called robust learning. In the literature, it is possible to find a series of recent works concerned with developing uncertainty-aware models \citep{Das2021,Psaros}.

Uncertainty is an important topic as developing robust models is essential for applying an ML in a real-case scenario. Most application domains have associated uncertainties caused by corrupted data, measurement noise, redundancies, and instrument uncertainties. When these issues are not considered, the ML tools may perform poorly and become inadequate. This might create a general skepticism in the general scientific community towards ML tools. Hence, robust learning plays an essential role in SciML literature.

Despite its increasing interest, there is still a lack of algorithms that efficiently cope with the epistemic and aleatory uncertainties in real case scenarios \cite{Das2021,Li2018}.  For instance, \citet{Gal2015} present a seminal work regarding this topic. The authors proposed evaluating reinforcement learning models using a Bayesian approximation technique. The referred article points toward issues there still needs to be addressed in this field. Some of these issues will be addressed in the present work.

Fully understanding the uncertainty of ML models is still a complex issue as it means evaluating the uncertainty of predictions. \citet{Abdar2021} present a comprehensive review on this topic and reinforce the ideal of an increasing necessity for further developments. \citet{Costa2022} proposed a novel strategy for developing uncertainty-aware soft sensors based on Deep Learning architecture. The authors mentioned the need for developing methods that simultaneously evaluate the different uncertainty components involved in an ML model identification.  

While identifying an ML model, there are several perspectives to be considered. For instance, \citet{Abdar2021} proposed three primary uncertainties related to an ML application, the absence of theory and causal models, the sensitiveness to data corruption or imperfection, and the computational effort.  As mentioned, several methodologies in the literature address one of these listed points. However, the literature lacks methods that can address all of them in the same framework.

Hence, this work proposes a comprehensive methodology for uncertainty evaluation of the ML model for SciML. The proposed method considers the epistemic and aleatory uncertainties related to both data used to train the models and the model itself. Therefore, it is possible to provide an overall strategy for the uncertainty-aware models in the SciML field. This work contributes to the robust learning literature as it allows for an approach that considers the several sources of uncertainties involved in the SciML model identification. The proposed methodology uses Bayesian inference to evaluate non-linear models' uncertainty propagation for ML models through Monte Carlo methods.

\section{Methodology Monte Carlo uncertainty training}
\label{SEC:01}

\begin{figure}
    \centering
    \includegraphics[width=\columnwidth]{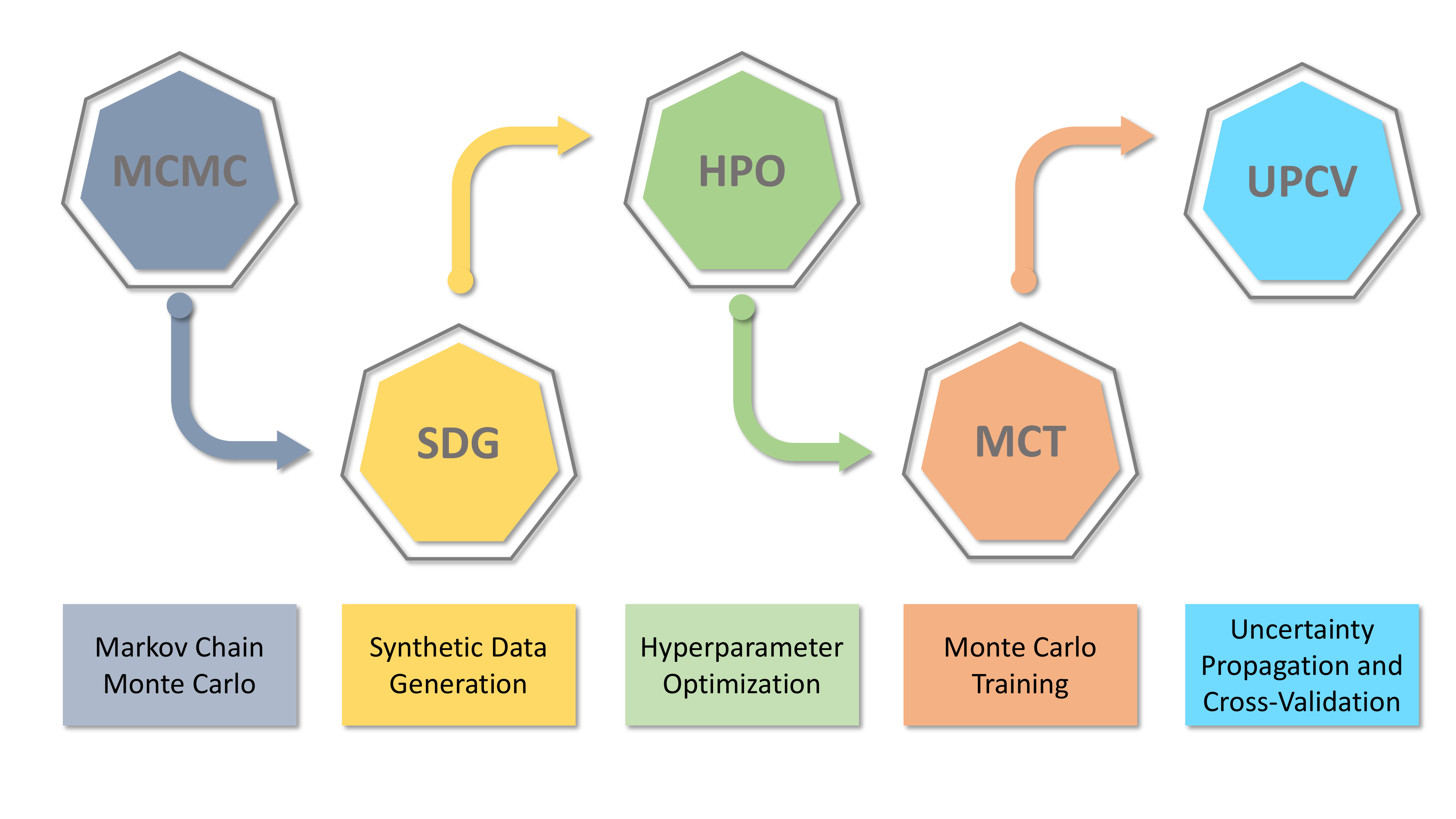}
    \caption{General methodology scheme.}
    \label{fig:01}
\end{figure}

This paper proposes a general methodology to build models based on artificial intelligence (AI) with uncertainty assessment. This proposed methodology is divided into five steps to obtain the validated machine learning models with uncertainty assessment. Fig. \ref{fig:01} presents the general scheme of the proposed method as follows.

The first step is using the Markov Chain Monte Carlo \cite{Gamerman2006} method to obtain the uncertainty of the non-linear model parameters that represent the system. Following the methodology, the validated model is used to generate synthetic data. This synthetic data is used to build the neural networks in two steps. The first step is to define the type and the general architecture of the neural network before optimizing the hyperparameters. After finding the best network architecture, the second step is to perform a Monte Carlo simulation training to propagate the uncertainty from the non-linear model (i.e., synthetic data) to the AI model. The last step of the proposed methodology is to perform the validation and uncertainty assessment of the trained model. In that step, the AI model prediction and simulation are collated with the non-linear model and experimental data if it is available. Also, the uncertainty assessment is performed in that step. The following subsections will provide specific details about each stage of the proposed methodology.

\subsection{Markov Chain Monte Carlo}
\label{SUBSEC:2_1}

Phenomenological and empirical models have common parameters that interfere with their respective output behavior. In these models, parameter estimation is a challenge because their choice contributes to the prediction uncertainty of the model. Consider a non-linear dynamic model written as:
\begin{equation}
    \dot{\mathbf{y}} = \mathbf{f}(t,\mathbf{x},\mathbf{u},\boldsymbol{\theta}),
    \label{EQ:01}
\end{equation}
\noindent where $\mathbf{f}$ is the relationship function between the time $t$, states $\mathbf{x}$, the input vector $\mathbf{u}$, parameters $\boldsymbol{\theta}$, and the output vector $\mathbf{y}$. In this scenario, if $\boldsymbol{\theta}=[\theta_1,\theta_2,…,\theta_{np}]$ is a set of $np$ parameters, and if they have a low predictive probability, the model will not provide a good forecast or adjustment to experimental data \cite{Migon2014}. Thence, it is essential to know the model's parameters' probability density function (PDF) and, consequently, the model uncertainty to ensure that the model and the parameters are good enough.

Several methods are available in the literature to solve the inference problem, estimate parameters and the associated uncertainty. \citet{Bard1974} presents several tools to obtain the variance of models from the frequentist approach, including Least Squares and Maximum Likelihood methods. The main drawback of \citet{Bard1974} methods is the hypothesis imposed to obtain the variance of the parameters, including Gaussian distribution. However, the Bayesian approach estimates the joint probability distribution using all available information about the system without assumptions about the target distribution \cite{Gamerman2006}.

The Bayesian approach to inference allows obtaining the posterior PDF of any set of parameters $\left(g_\theta \left(\eta \mid D,I\right)\right)$ using with information the observation data $(D)$ and any previous information about the system $(I)$. Therefore, using the Bayes Theorem, it is possible to write the following relationship between the earlier variables \cite{Gamerman2006}:
\begin{align}
    g_{\boldsymbol{\theta}} (\eta\mid D,I)\propto L(\eta\mid D) g_{\boldsymbol{\theta}} (\eta\mid I),
    \label{EQ:02}
\end{align}
\noindent where $\eta$ represents sampled values of $\boldsymbol{\theta}$, $L$ is the likelihood function, and $g_{\boldsymbol{\theta}} (\eta\mid I)$ is the prior distribution of $\boldsymbol{\theta}$, that is a new observation of $\boldsymbol{\theta}$. Eq. \ref{EQ:02} allows updating the actual knowledge of the system represented by the posterior $g_{\boldsymbol{\theta}} (\eta\mid D,I)$.

The likelihood $L(\eta\mid D)$  is defined by \citet{Migon2014} as a function that associates the value of the probability $g_\theta (\eta \mid I)$ with each $\eta$ value. By defining the estimation process as a least square problem, the objective will be to minimize a loss function that can be represented by a weighted least square estimator (WLSE). So, the likelihood function can be defined as:
\begin{align}
L(\eta \mid D)\propto-\frac{1}{2} \sum_{i=1}^n \left( y_i^{exp}-y_i^m \right)^{\top} \boldsymbol{\Phi}^{-1} \left( y_i^{exp}-y_i^m \right), 
\label{EQ:03}
\end{align}
\noindent where $(y_i^{exp}-y_i^m)$ is the residual between the experimental data $(y_i^{exp})$ and the obtained with the prediction model $(y_i^m)$. Also, the use of the WLSE estimator implies the use of the variance of the residual between the $y^{exp}$ and $y^m$ for the $ny$ outputs of the system, represented in Eq. \ref{EQ:03} as $\boldsymbol{\Phi}=diag[\Phi_1,\Phi_2,…,\Phi_{ny}]$.

The posterior PDF of each $\theta_i$ parameter of the vector $\boldsymbol{\theta}$, $g_\theta \left(\eta \mid D,I\right)$ is obtained found the marginal posterior density function $g\left(\theta_1,\theta_2,…,\theta_{np}\right)$ and is defined by \citet{Gamerman2006} as:
\begin{align}
    g_\theta (\eta \mid D,I) \propto \int_\theta \left( L(\eta \mid D) g_\theta \right) (\eta \mid I)d\theta_{n-j}
    \label{EQ:04}
\end{align}

\begin{figure}
    \centering
    \includegraphics[width=0.5\columnwidth]{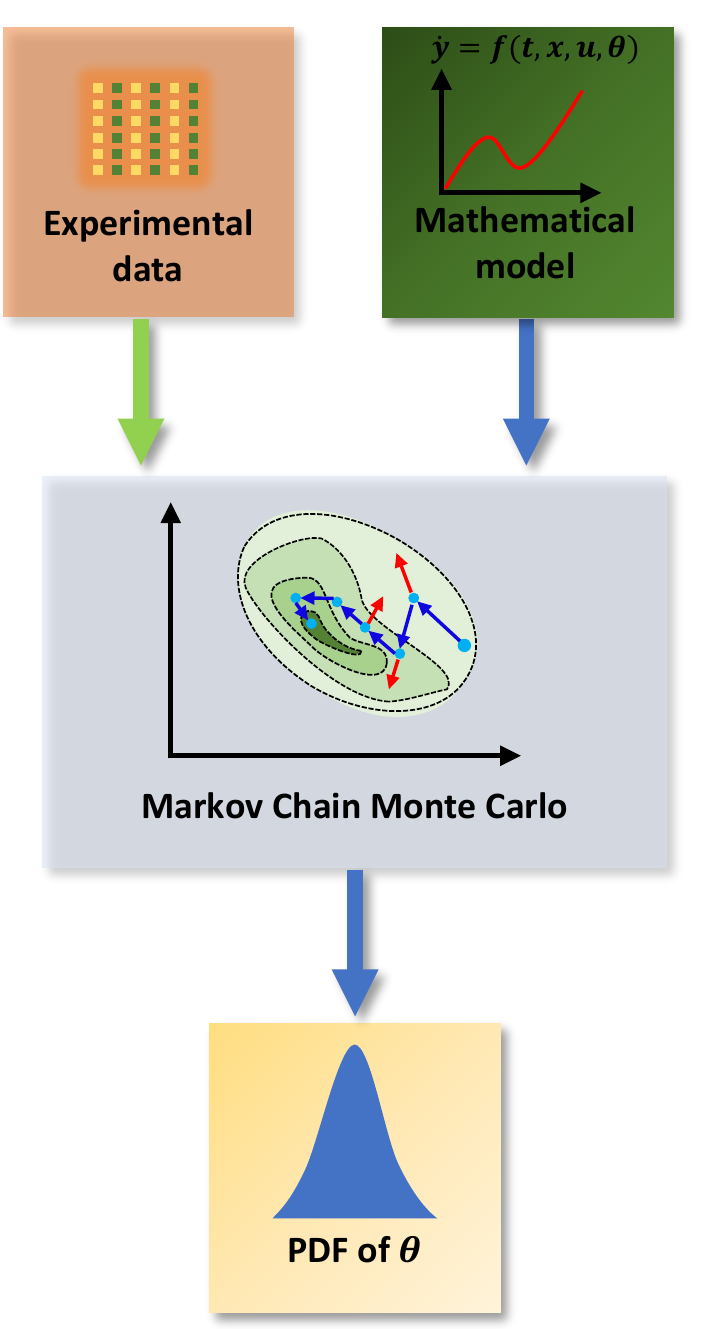}
    \caption{MCMC method.}
    \label{fig:02}
\end{figure}

Chapter 5 by \citet{Gamerman2006} presents several methods for solving the inference problem of Eq. \ref{EQ:04}. The Markov Chain Monte Carlo methods have some interesting features among the numerical integration methods. Among these characteristics is convergence because when chains are adequately constructed, after a sufficiently high number of iterations, the chains will converge to an equilibrium distribution. Thus, Fig. \ref{fig:02} shows a schematic diagram of the solution to the inference problem using the MCMC method. The general idea is that the MCMC uses existing information, such as experimental data and a likelihood function, to provide a mathematical model and the associated PDF of the estimated parameters.

On the other hand, this paper proposes using the DRAM (Delayed Rejection Adaptive Metropolis) MCMC algorithm \citet{Haario2006} presented to solve the inference problem. The DRAM algorithm combines the Adaptive Metropolis, which provides global adaptation, and the Delayed Rejection, which offers local adaptation. Also, the main idea of the DRAM algorithm is to collect information during the chain run and tune the target PDF by using the learned information.

\subsection{Synthetic data generation}
\label{SUBSEC:2_2}

In building AI models, data quality is crucial to obtain great-adjusted models. This paper proposes a methodology to build AI models using synthetic data from a non-linear model. Also, the methodology is based on the law of propagation of uncertainty established in \citet{BIPM2008,BIPM2008b,BIPM2011}. In this sense, the quantity and quality of data used in training must be adequate. Non-linear models in optimization and control applications often represent a high computational cost. These computational efforts can be reduced using artificial intelligence models trained with data from validated non-linear models.

\begin{figure}
    \centering
    \includegraphics[width=0.5\columnwidth]{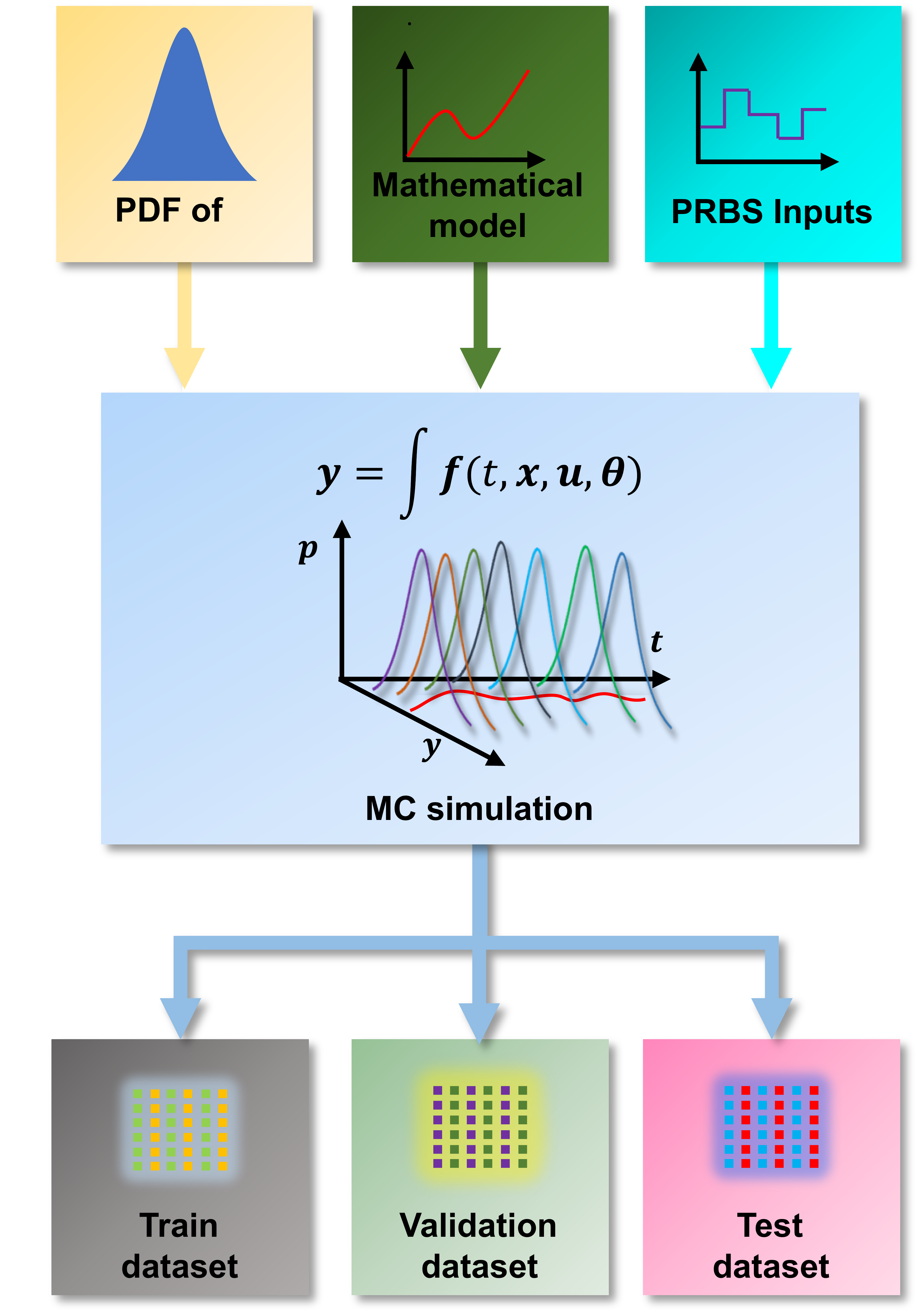}
    \caption{Monte Carlo simulation method to data generating.}
    \label{fig:03}
\end{figure}

This paper proposes to build the training database by drawing a sample of the parameters PDF and propagating it to the outputs of the non-linear model. In this way, the data used for training the models need to be representative of the operating conditions of the system and characterize the uncertainties of the non-linear model. Fig. \ref{fig:03} presents a generic Monte Carlo Method (MCM) simulation scheme for synthetic data generation. Fig. \ref{fig:03} scheme is inspired by the algorithm for implementing the Monte Carlo Method presented in Supplement 1 and 2 to the "Guide to the expression of uncertainty in measurement" \cite{BIPM2008,BIPM2011}.

The first step of the algorithm is to determine the number $m$ of trials to be performed. In general, the MCM will produce better responses the greater the number of shots, and the recommendation is to use a value $m=10^6$ \cite{Migon2014}. However, smaller values can be used when evaluating complex models requiring high computational costs for numerical solutions. This reduction in the number of tests may not allow the correct characterization of the PDF of the output values and produce less reliable results \cite{BIPM2008}.

After defining the number of tests to be performed, a matrix $\Theta$ containing $m$ vectors of PDF samples of $g_\theta (\eta  \mid D,I)$ is built:
\begin{equation}
\boldsymbol{\Theta} = 
\begin{pmatrix}
\boldsymbol{\theta}_{1,1} & \dots & \boldsymbol{\theta}_{1,np}\\
\vdots & \ddots & \vdots\\
\boldsymbol{\theta}_{m,1} & \dots & \boldsymbol{\theta}_{m,np}
\end{pmatrix}.
\label{EQ:05}
\end{equation}

Thus, $m$ distinct set of model parameters are obtained so that it is possible to write
\begin{equation}
    \dot{\mathbf{y}}_{1, \dots, m} = \mathbf{f}(t,\mathbf{f},\mathbf{u},\boldsymbol{\Theta} ),
\label{EQ:06}    
\end{equation}

\noindent in this way, it is possible to integrate the model so that the $m$ dynamic responses are obtained for the output variables $y$.

As it is a dynamic model, it is necessary to obtain representative data from the entire operating region. An independent Pseudo-Random Binary Sequence (PRBS) signal for each of the u inputs is generated through a Latin Hypercube Sampler (LHS). The PRBS signals are then combined with the non-linear mathematical model of the system. A dynamic response is received for each parameter combination obtained from the PDF of the parameters. Additionally, the system's dynamic response will be represented by a set of curves obtained from this MC simulation. In this scenario of propagation of uncertainties by MCM in dynamical systems, the true value must be calculated for each sampling instant in which the equations were solved. In this way, it is considered that for each sampling instant, a PDF sample is obtained for each system output response $\mathbf{y}(t)$.

From Fig. \ref{fig:03}, it is also possible to observe that the data generated through the MC simulation will be divided into three sets: training, testing, and validation. This Division follows the common literature guidelines for training AI models \cite{Haykin1999}. The first two sets are used during supervised training of the models since the algorithms need two data sets for training and validation. The third set, the test dataset, is used for final cross-validation so that these data are "unknown" to the AI model. Thus, the ability to predict and extrapolate to new data is evaluated. Additionally, dividing the data into three sets must consider that the training process has consistent information. In this sense, sets are typically divided into: Train - 70\%, Validation - 15\% and Test - 15\%.

\subsection{Data Curation}
\label{SUBSEC:2_3}

A database to train a dynamic data-driven model should be systematically organized to represents the system dynamics. Hence, the identified model can approximate the observed dynamic phenomena. There are several ways to manage a data set to incorporate the system's time dependence. The most common is Non-linear autoregressive with exogenous inputs (NARX) predictors. The general NARX structure is composed of a prediction  of the actual output  as:
\begin{equation}
    y_k=f(x_k,x_{(k-1)},x_{(k-2)},…,x_{(k-n)},y_{(k-1)},y_{(k-2)},…,y_{(k-p)},\gamma)+\epsilon
    \label{EQ:07}
\end{equation}

\noindent where $x_{(k-1)},  x_{(k-2)},\dots, x_{(k-n)}$ is the input delay, and $y_{(k-1)},y_{(k-2)},…,y_{(k-p)}$   are the $p$ of past values for the output and input, respectively. The noise, $\epsilon$, is additive: for the NARX, the error information is assumed to be filtered through the system's dynamic. In it turn, $\gamma$ is an parameter that represents the model parameters uncertainty. As observed from Eq. \ref{EQ:07}, the NARX predictor has two hyperparameters: $n$ and $p$. These parameters should be defined correctly to improve the dynamic representativeness of the data. For this purpose, \citet{He1993} proposed the Lipschitz coefficient analysis. A Lipschitz coefficient is then calculated for each pair of measurements. For further information about the Lipschitz coefficients calculations, see \citet{He1993}.

\subsection{Building the AI model}
\label{SUBSEC:2_4}

\begin{figure}
    \centering
    \includegraphics[width=0.5\columnwidth]{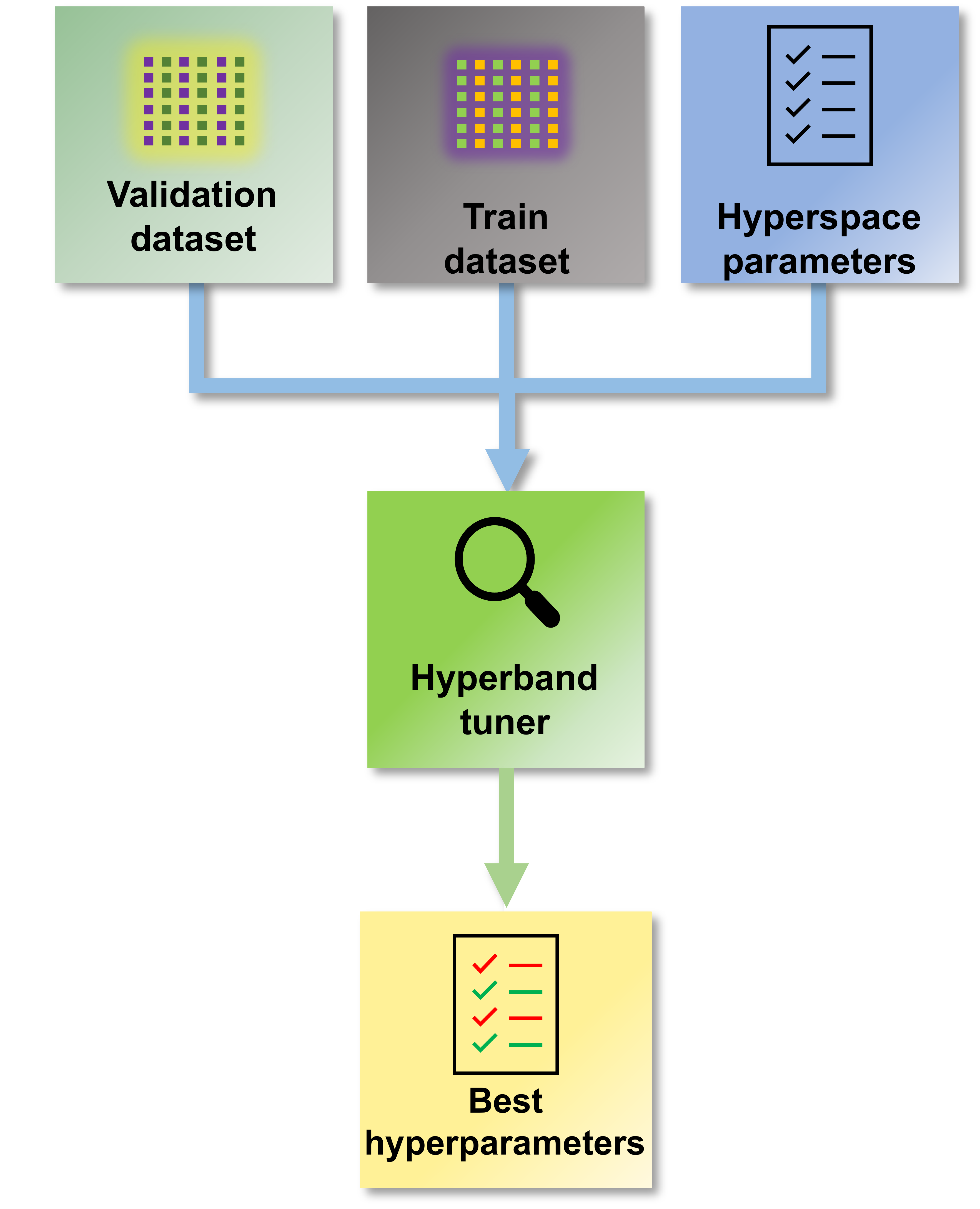}
    \caption{Optimization procedure to find the hyperparameters of the neural network.}
    \label{fig:04}
\end{figure}

Fig. \ref{fig:04} shows a methodology step responsible to find hyperparameters of the network. That step defines the appropriate architecture and network format that one wants to get before starting the training. The type of network is the first aspect to be evaluated when building an AI model. In this way, expert knowledge must be considered to define whether a network will be used, e.g., recurrent, convolutional, or dense. Choosing the network format depends on the characteristic of the system being modeled and what is the main application of the model.

Once the general format of the type of network is defined, the next step is to define its architecture. There are some powerful algorithms available in the literature for this means. The optimal number of layers determines the architecture of a neural network, the number of neurons per layer, and activation functions, among others \cite{Li2016}. Determining these parameters is one source of uncertainties during the modeling process that needs to be considered.

\subsection{Monte Carlo training}
\label{SUBSEC:2_5}

The fourth step of the methodology proposed in this work is the uncertain training of networks through Monte Carlo simulations. The stages of generating data, obtaining the architecture of the neural network, and the hyperparameters provide the necessary information for the uncertain training of the networks. In this way, it is sought to characterize the prediction region of the identified non-linear system, obtained in Sec. \ref{SUBSEC:2_1}, through a set of networks capable of representing each of the probable outputs of the model.

\begin{figure}
    \centering
    \includegraphics[width=0.5\columnwidth]{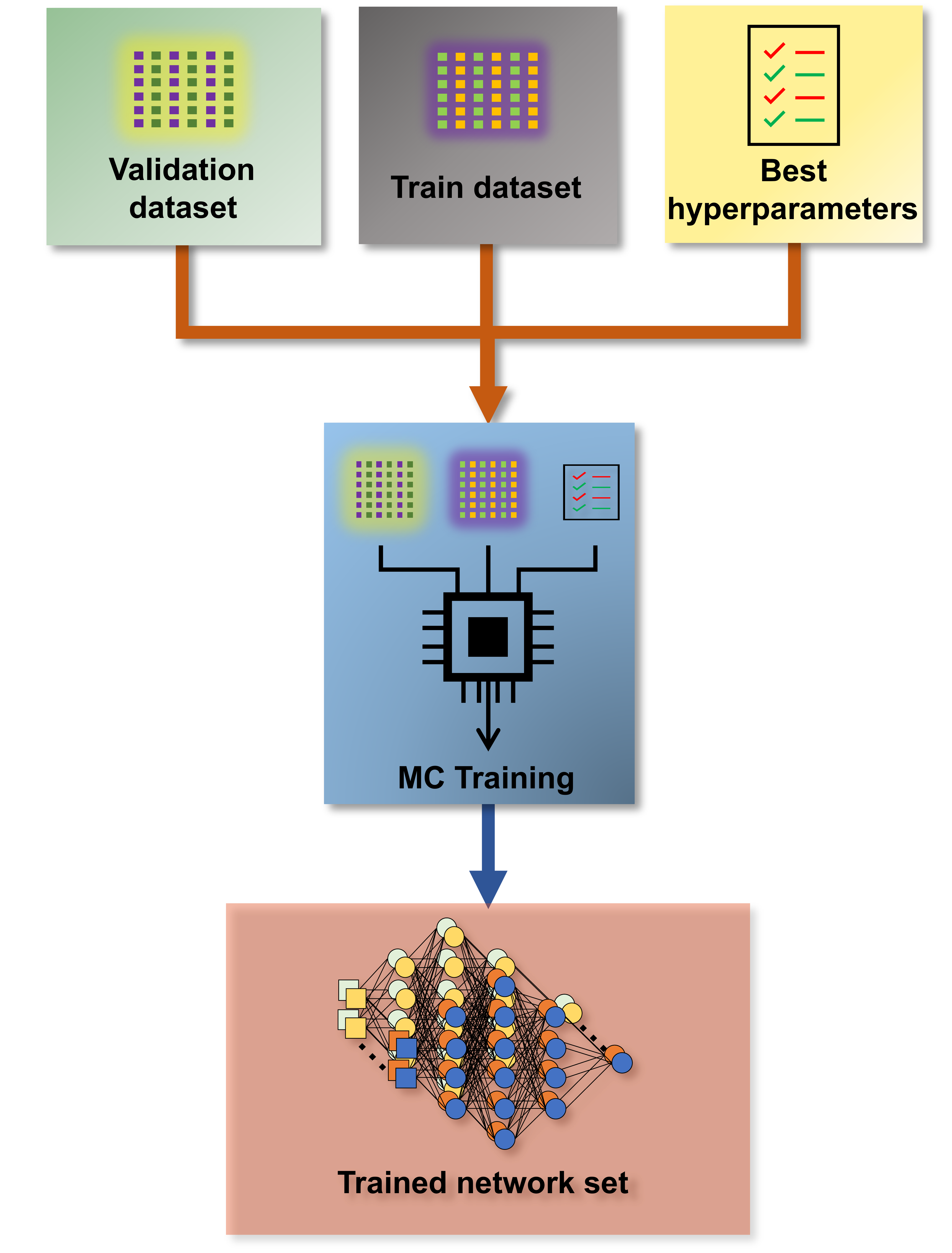}
    \caption{Monte Carlo training method.}
    \label{fig:05}
\end{figure}

Fig. \ref{fig:05} presents a simplified schematic diagram of this step in the methodology. It is possible to observe that the MC Training stage boils down to massive training on top of the generated data. The result is a set of equally probable trained AI models that comprehensively account for the uncertainty sources.

This training step follows the concepts of Monte Carlo simulations. It is the step that has the most significant computational effort. The simulation is built to train an AI model with pre-defined optimal architecture for each element in the training, validation, and test datasets. In addition to the computational effort required to perform this step, the volume of data generated will also be significant. However, the amount of data generated has a less relevant impact.

On the other hand, nowadays, SciML model training has become increasingly efficient. The literature has presented algorithms that extract maximum performance from the available hardware. Additionally, manufacturers have built hardware with specific characteristics for application training and execution of Artificial Intelligence. In this scenario, the available technology makes it feasible to run a Monte Carlo simulation for training neural networks.

\subsection{Propagation and cross-validation}
\label{SUBSEC:2_6}

The methodology proposed in this article includes a supplementary validation stage of the built AI model. In this step, the data used for cross-validation are labeled “Test data” in Fig. \ref{fig:03} (Note that the set called "validation" is used during training.). In a complementary way, the validation in the context of dynamic models with uncertainty needs to be evaluated to compare the coverage regions. In this context, a model is considered validated when the coverage of regions of models is overlapping, implying that the values are statistically equal.

The main issue involved is the method used to assess the uncertainty of the model. For both the non-linear phenomenological model and the AI model, this article proposes to use the Monte Carlo method. Thus, the uncertainty of the evaluated model can be obtained assuming the same hypotheses proposed by \citet{Haario2001,Haario2006}; That is: the variance is approximated by an inverse Gamma distribution. Then:
\begin{equation}
    V[y] \approx \Gamma^{-1}(x,\alpha,\beta)
    \label{EQ:08}
\end{equation}
\noindent where the distribution is supported in $x>0$ and represented by $\Gamma^{-1}$ and the $\alpha$ and $\beta$ parameters are the shape and scale of the distribution.

In it turn, to obtain the parameters, $\alpha$ and $\beta$, \citet{Gelman2013} suggest using:
\begin{equation}
    \alpha(j)=\frac{N_{prior} (j)+N_{data} (j)}{2},
    \label{EQ:09}
\end{equation}
\begin{equation}
    \beta(j) = \frac{2}{N_{prior}(j)\cdot V_0^2 + SSE(j)},
    \label{EQ:10}
\end{equation}

\noindent where $j=1,2,…,ny$ is the number of outputs of the model. On the other hand, in calculating $V_0^2$  is the variance of Prior, and $SSE(j)$ is the sum of the squared errors between the prediction and the experimental data.

Using the hypothesis of non-informative prior, the variance of the prior and number of points are unknown. In this way, the previous equations can be approximated by:
\begin{equation}
    \alpha(j)=\frac{N_{data} (j)}{2},
    \label{EQ:11}
\end{equation}
\begin{equation}
    \beta(j) = \frac{2}{SSE(j)}.
    \label{EQ:12}
\end{equation}

The methodology of this article implies obtaining two sets of model parameters that will each have their associated uncertainty. Thus, the proposal is that the variance calculation is performed using the SSE obtained through the estimation data when the MCMC is performed. The SSE is obtained with the training data for the training of networks. Then, the uncertainty of a prediction will be based on the variance of the model that will follow the inverse gamma distribution. This methodology allows for characterizing the epistemic uncertainty of the model.

\section{Results and discussion}
\label{SEC:3}
This section presents the results of applying the proposed methodology in a case study. A polymerization reactor with synthetic data was used as a case study. The detailed polymerization model is presented in Sec. \ref{SUBSEC:3_1}. Next, it explains networks' construction using the Monte Carlo method, propagation, and final cross-validation, following the proposed methodology.

\subsection{Case study: Polymerization reactor}
\label{SUBSEC:3_1}

The reactor model is presented in detail by \citet{Alvarez2012}. It is composed by a system of algebraic differential equations (DAE) with thirteen equations, as follows:
\begin{align}
    &\frac{d[I]}{dt} = \frac{Q_i [i_f] - Q_t [I]}{V} - k_d [I], \label{EQ:13}\\
    &\frac{d[M]}{dt} = \frac{Q_m [M_f] - Q_t [M]}{V} - k_p [M][P], \label{EQ:14}\\
    &\frac{dT}{dt} = \frac{Q_i[T_f-T]}{V} +\frac{-\Delta H_r}{\rho C_p}k_p [M][P] - \frac{hA}{\rho C_p V}(T-T_c), \label{EQ:15} \\
    &\frac{dT_c}{dt}= \frac{Q_c (T_{cf}-T_c)}{V_c} + \frac{hA}{\rho C_p V}(T-T_c), \label{EQ:16}\\
    &\frac{dD_0}{dt}= 0.5k_t [P]^2  - \frac{Q_t D_0}{V}, \label{EQ:17}\\
    &\frac{dD_1}{dt}= M_m k_p [M][P]  - \frac{Q_t D_1}{V}, \label{EQ:18}\\
    &\frac{dD_2}{dt}= 5M_m k_p [M][P] +M_m \frac{k_p^2}{k_t}[M]^2 - \frac{Q_t D_2}{V}, \label{EQ:19}\\
    &[P] = \left[ \frac{2f_i k_d[I]}{k_t} \right]^{0.5}, \label{EQ:20}\\
    &Q_t = Q_i + Q_s + Q_m, \label{EQ:21}\\
    &\Bar{M}_w = M_m \frac{D_2}{D_1},\label{EQ:22}\\
    &PD = M_m \frac{D_2 D_0}{D_1^2}, \label{EQ:23}\\
    &\eta = 0.0012(\Bar{M}_w)^{0.71}. \label{EQ:24}
\end{align}

In the above DAE system, Eq \ref{EQ:13} to \ref{EQ:16} represent the mass and energy balance of the Monomer and Initiator. Eq. \ref{EQ:17} to \ref{EQ:19} are the moments' equations of the dead polymer, in which $D_0$, $D_1$, and $D_2$ represents the moments of the dead polymer. The algebraic equations are used to describe the relationship between supplementary variables. Eq. \ref{EQ:22} represents the weight-average molecular weight, and Eq. \ref{EQ:24} represents the viscosity. Tab. \ref{tbl1}, \ref{tbl2} and \ref{tbl3} show the model's parameters' general definition, the initial condition value of inputs and steady-state of outputs systems variables.

\citet{Alvarez2012} developed this model based on seven hypothesis which are: the lifetime of the radical polymer is shorter than other species; Long Chain Assumption (LCA) related to the monomer consumption; the chain transfer reaction to monomer and solvent can be neglected; operation below 373K because greater temperatures cause monomer thermal initiation;  termination by disproportionation is not considered; the rate of termination is dominant; only the heat of polymerization is considered.

\begin{table}[width=.9\linewidth,cols=4,pos=h]
\caption{Parameters and initial conditions.}\label{tbl1}
\begin{tabular*}{\tblwidth}{@{} LL@{} }
\toprule
Nominal Process Parameters &	Value \\
\midrule
Frequency factor for initiator decomposition, $A_d  (h^{-1})$ &	$2.142\times10^{17}$\\
Activation energy for initiator decomposition, $E_d  (K)$	& $14897$\\
Frequency factor for propagation reaction, $A_p(L\cdot mol^{-1}\cdot h^{-1}  $ & $3.81\times10^{10}$\\
Activation temperature for propagation reaction, $E_p  (K)$ &	$3557$\\
Frequency factor for termination reaction, $A_t  (Lmol^{-1}h^{-1})$&	$4.50\times10^{12}$\\
Activation temperature for termination reaction, $E_t  (K)$&	$843$\\
Initiator efficiency, $f_i$ & 	$0.6$\\
Heat of polymerization, $-\Delta H_r  (J\cdot mol^{-1})$&	$6.99\times10^4$\\
Overall heat transfer coefficient, $hA (J\cdot K^{-1} \cdot L^{-1})$	 & $1.05\times10^6$\\
Mean heat capacity of reactor fluid, $\rho C_p  (JK^{-1}L^{-1})$ &	$1506$\\
Heat capacity of cooling jacket fluid, $\rho_c C_pc  (JK^{-1}L^{-1})$ &	$4043$\\
Molecular weight of the monomer, $M_m  (g\cdot mol^{-1})$&	$104.14$\\
\midrule
Initial conditions &	Value\\
\midrule
Reactor volume, $V(L)$ & $3000$\\
Volume of cooling jacket fluid, $Vc (L) $ & $	3312.4 $\\
Concentration of initiator in feed, $I_f (mol \cdot L^{-1})$&	$0.5888$\\
Concentration of monomer in feed, $ M_f (mol\cdot L^{-1})$&	$8.6981$\\
Temperature of reactor feed, $T_f (K) $ &	$330 $ \\
Inlet temperature of cooling jacket fluid, $ Tcf(K)$&	$295$\\
\bottomrule
\end{tabular*}
\end{table}

\citet{Alvarez2012} discuss the Polymerization reactor from the control and optimization point of view. However, other relevant aspects are pointed out. One of these, \citet{Alvarez2012} uses the Eq. \ref{EQ:23} as a virtual analyzer for the viscosity because this is a difficult variable to measure in the studied reactor. The authors use the temperature and the viscosity as controlled variables manipulating the initiator flow rate and the rate of the cooling jacket. Therefore, this case study used to validate the proposed methodology for uncertainty assessment of neural networks. Also, using an AI model reduces the computational efforts of the control and optimization loops.

As no experimental data is available regarding this system, this paper proposes using random white noise to simulate the interferences that usually occur in an experimental setup. The system was simulated with the initial and steady-state conditions in Tab. \ref{tbl2} and \ref{tbl3}. 

\begin{table}[width=.9\linewidth,cols=4,pos=h]
\caption{Steady-state inputs conditions and LHS region.}\label{tbl2}
\begin{tabular*}{\tblwidth}{@{} LLLL@{} }
\toprule
Variable & Steady-state  & Minimum &	Maximum \\
\midrule
Flow rate of initiator, $Q_i  (L\cdot h^{-1})$ &	108 &	91.8 &	124.2 \\
Flow rate of solvent, $Q_s  (L\cdot h^{-1})$ &	3312.4 &	2815.5 &	3809.26 \\
Flow rate of monomer, $Q_m  (L\cdot h^{-1})$ &	0.5888 &	0.5005 &	0.6771 \\
Flow rate of cooling jacket fluid, $Q_c  (L\cdot h^{-1}) $&	8.6981 &	7.3934 &	10.0028 \\
\bottomrule
\end{tabular*}
\end{table}

\begin{table}[width=.9\linewidth,cols=4,pos=h]
\caption{Output variables at steady-state.}\label{tbl3}
\begin{tabular*}{\tblwidth}{@{} LLLL@{} }
\toprule
Variable &	Value \\
\midrule
Concentration of initiator in the reactor, $I (mol.L^{-1})$ &	$330$ \\ 
Concentration of monomer in the reactor, $I (mol.L^{-1})$	& 295 \\
Temperature of the reactor, $T (K)$ &	$323.56$ \\ 
Temperature of cooling jacket fluid, $T (K)$	& $305.17$ \\ 
Molar concentration of dead polymer chains, $D_0  (mol.L^{-1})$	& $2.7547\times10^{-4}$ \\ 
Mass concentration of dead polymer chains, $D_1  (g.L^{-1})$	&$16.110$ \\
\bottomrule
\end{tabular*}
\end{table}

\begin{figure}
    \centering
    \includegraphics[width=0.8\columnwidth]{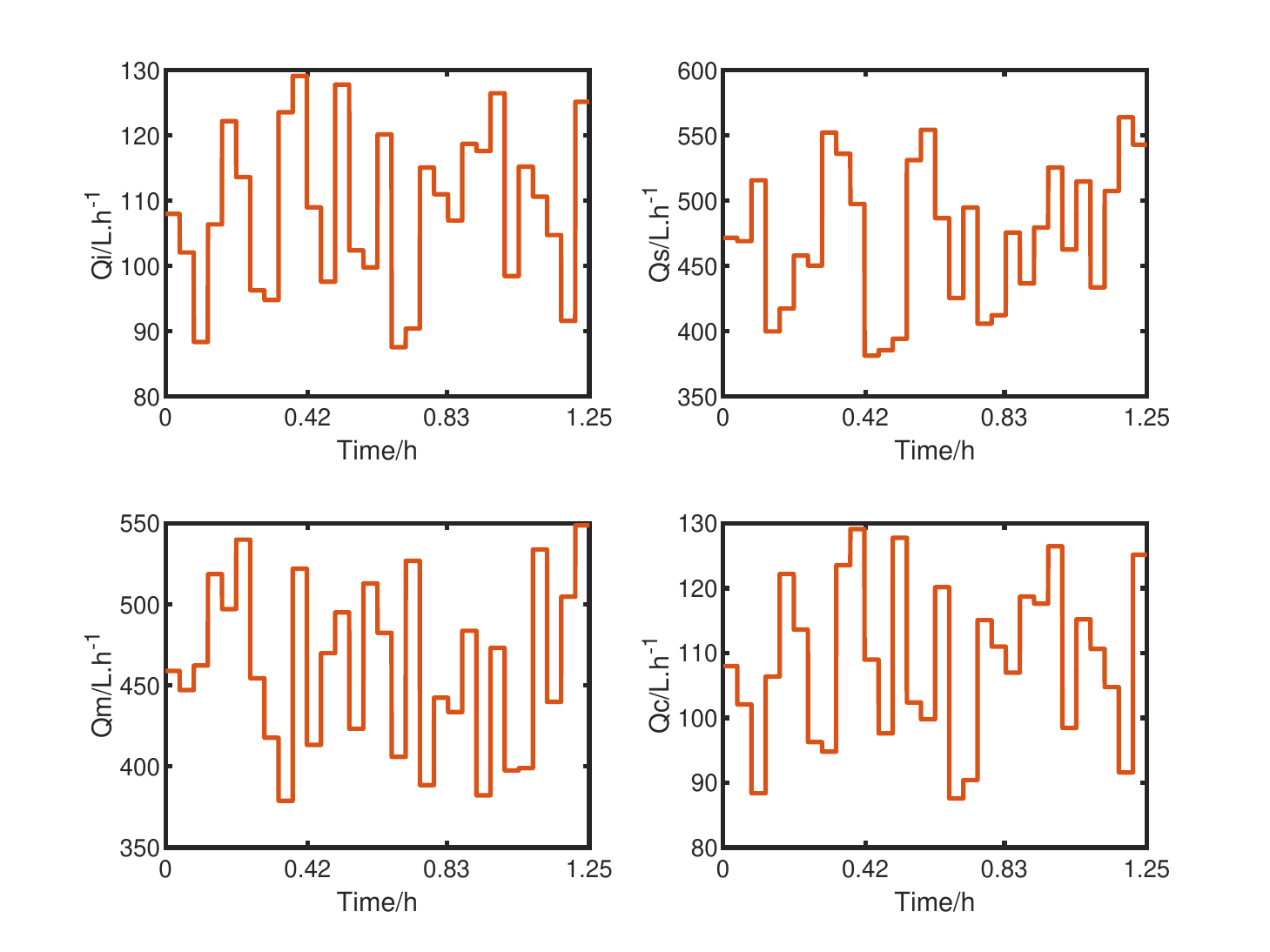}
    \caption{LHS inputs.}
    \label{fig:06}
\end{figure}

Fig. \ref{fig:06} shows the LHS generated with $\pm15$\% of the steady-state input value and used as input in the simulation. A total of 30 steps with 150 hours of simulation each were developed to compose the synthetic data.

\begin{figure}
    \centering
    \includegraphics[width=0.58\columnwidth]{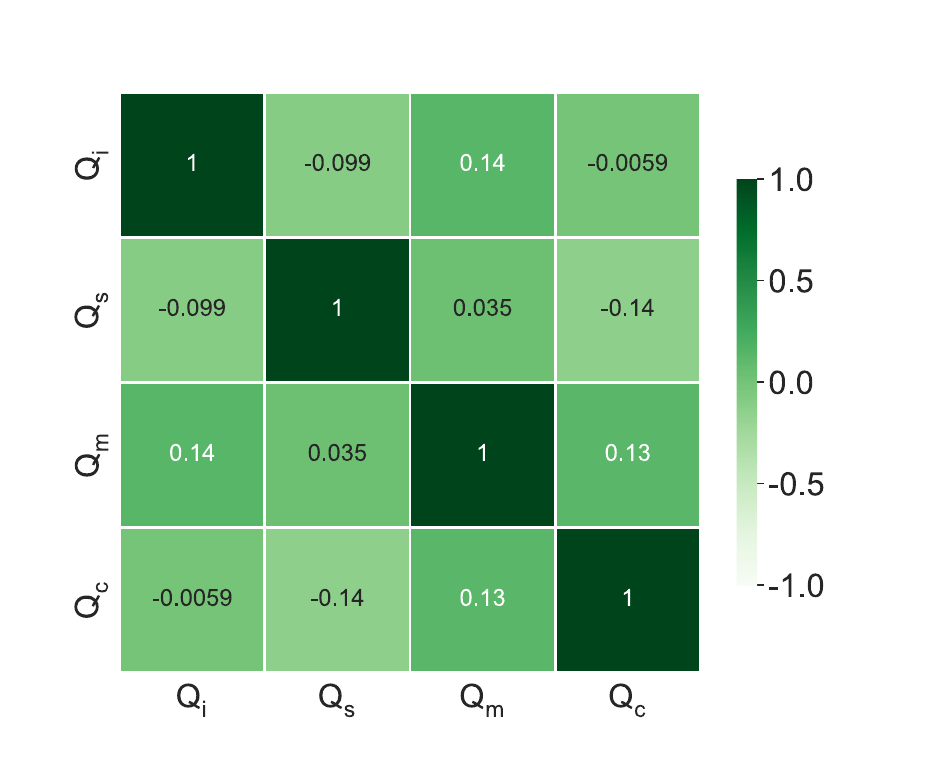}
    \caption{Inputs correlation map.}
    \label{fig:07}
\end{figure}

Fig. \ref{fig:07} presents the correlation heat map for the variables set by the LHS. The main diagonal presents high correlations as it pairs the variables with themselves. In contrast, the pairing between the other entries produces a value that allows verifying if the generated values are uncorrelated. As low as these values are, the less correlated, they will be. As shown in Fig. \ref{fig:07}, the correlations are close to null, demonstrating that the LHS can efficiently generate uncorrelated samples.

\begin{figure}
    \centering
    \includegraphics[width=0.8\columnwidth]{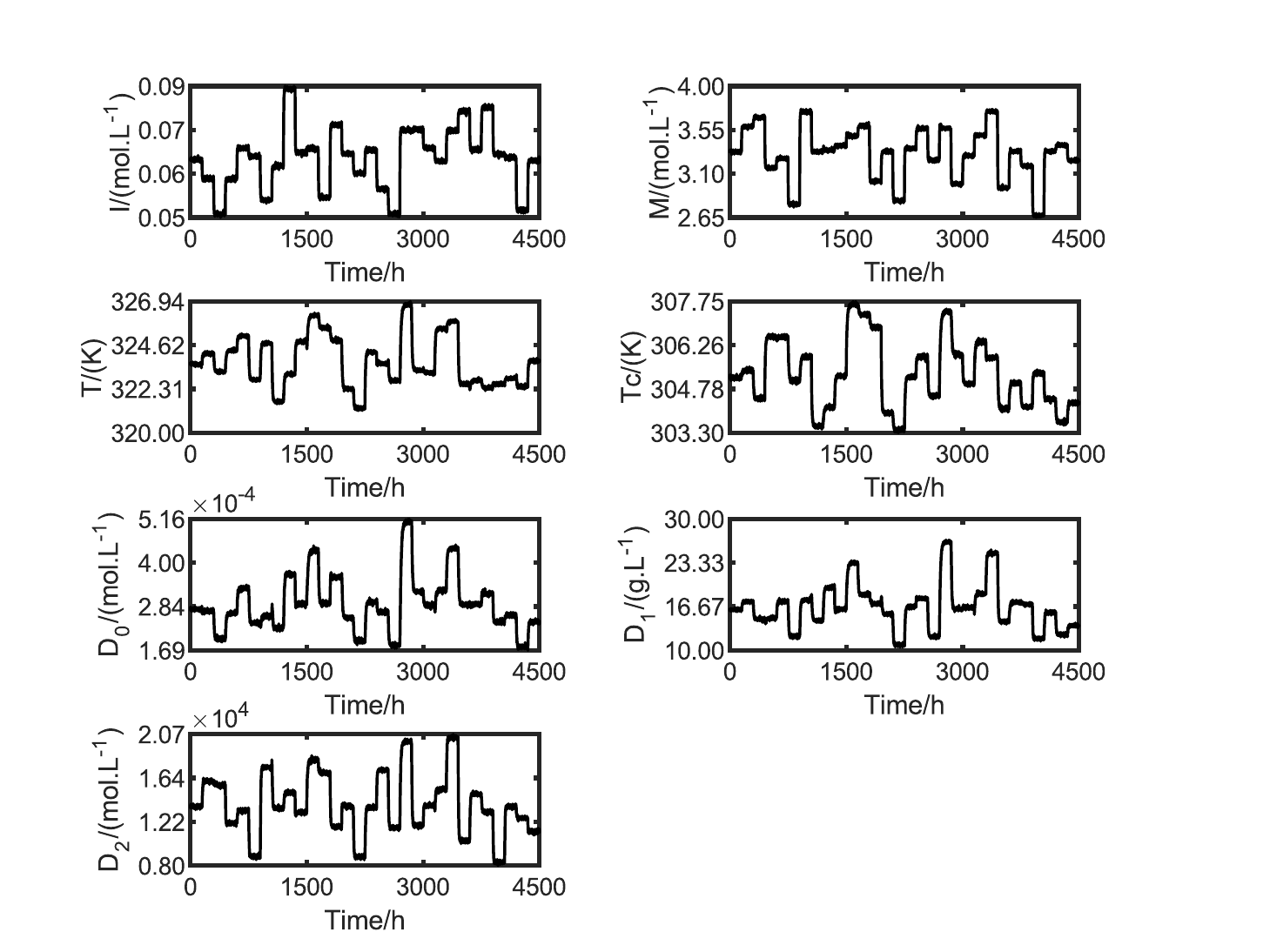}
    \caption{Synthetic output data with white noise.}
    \label{fig:08}
\end{figure}

The synthetic output dataset is shown in Fig. \ref{fig:08}. As mentioned, random noise with -20dB and 10\% of the output range is included in the signal to emulate the field conditions. All these variables are used in the Likelihood function presented in Eq. \ref{EQ:03}. 70\% of this data is used for the SciML model identification, and the rest is used for the methodology cross-validation step, as discussed in the subsequent sections.

\subsection{MCMC}
\label{SUBSEC:3_2}

The generated synthetic data allows the uncertainty assessment of the \citet{Alvarez2012} model, done through the MCMC methodology. Therefore, 12 model parameters and six initial conditions are used as decision variables for the MCMC algorithm. The DRAM algorithm proposed by \citet{Haario2006}  allows limiting the parameters search region. Therefore, in this paper, the search region was limited to $\pm 5$\% of the nominal value of the parameter. Also, all parameters were normalized to the nominal value to facilitate the algorithm's convergence.

Other algorithms aspects must be set. The first is the number of samples sorted to build the joint PDF. At this point, the algorithm was configured to build a set with 30 000 samples of the target joint PDF. The algorithm also was configured to use a non-informative prior. Therefore, to estimate the variance of the parameter, the algorithm chooses 5 000 samples to update the prior and discard the chain. After the burn step, the MCMC algorithm restarts to build the chains and evaluate the parameters in the search region.

Tab. \ref{tbl4} shows the resulting normalized parameters obtained from the Markov Chains. The MCMC algorithm does not make any assumptions about the target distribution. The mean and median in normal distributions converge to the same number. However, it is more conservative to assume that the distribution is not gaussian and use the median as a value for the most probable value. In Tab. \ref{tbl4}  is possible to see the difference between the parameters mean and median. Tab. \ref{tbl4}  also shows the standard deviation (std) and the geweke diagnose parameter \cite{Brooks}. For some parameters, the std value is relatively high. However, the geweke parameter indicates that the chain converges \cite{Brooks}. Figs. \ref{fig:Coverage_regions_1} to \ref{fig:Random_Walk_1} in the supplementary material (Appendix \ref{APPEND}) shows the confidence region and the fully Markov Chain. In those figures, it is compared a gaussian region and an unshaped region proposed by \citet{Possolo2010}.

\begin{table}[width=.9\linewidth,cols=4,pos=h]
\caption{Normalized parameters obtained by DRAM algorithm.}\label{tbl4}
\begin{tabular*}{\tblwidth}{@{} LLLLL@{} }
\toprule
Parameter  & Mean                           & Median                        & STD                           & Geweke                        \\
\midrule
$A_d $      & $1.00\times10^0 $    & $1.01\times10^0 $   & $1.96\times10^{-2}$ & $9.95\times10^{-1}$ \\
$E_d $      & $1.00 \times10^0 $    & $1.00\times10^0$    & $6.97\times10^{-4}$ & $9.99\times10^{-1}$ \\
$A_p $      & $9.99\times10^{-1}$  & $9.98\times10^{-1}$ & $1.99\times10^{-2}$ & $9.96\times10^{-1}$ \\
$E_p $      & $9.98\times10^{-1}$  & $9.98\times10^{-1}$ & $3.17\times10^{-3}$ & $9.99\times10^{-1}$ \\
$A_t  $     & $1.00\times10^0 $    & $1.00\times10^0   $ & $2.06\times10^{-2}$ & $9.83\times10^{-1}$ \\
$E_t  $     & $9.91\times10^{-1}$  & $9.88\times10^{-1} $& $2.42\times10^{-2}$ & $9.91\times10^{-1}$ \\
$f_i  $     & $1.00 \times10^0 $   & $1.01\times10^0    $& $2.33\times10^{-2}$ & $9.94\times10^{-1}$ \\
$-\Delta H_r $    & $9.99\times10^{-1}$  & $1.00\times10^0$    & $7.93\times10^{-3}$ & $9.96\times10^{-1}$ \\
$hA    $     & $1.00\times10^0 $    & $1.00\times10^0    $& $8.11\times10^{-3} $& $9.96\times10^{-1}$ \\
$\rho C_p$      & $9.97 \times10^{-1}$ & $9.98\times10^{-1}$ & $9.80\times10^{-3}$ & $9.96\times10^{-1}$ \\
$\rho_c C_{pc}$ & $1.01 \times10^0 $   & $1.01\times10^0 $   &$ 9.63\times10^{-3} $& $9.97\times10^{-1}$ \\
$M_m   $    & $9.99 \times10^{-1}$ & $9.99\times10^{-1} $& $6.41\times10^{-4} $& $9.99\times10^{-1}$ \\
$V  $        & $9.98\times10^{-1}$  & $9.99\times10^{-1}$ &$ 3.71\times10^{-3}$ & $9.99\times10^{-1}$ \\
$Vc     $    & $1.00\times10^0$     & $1.00\times10^0   $ &$ 1.34\times10^{-2}$ & $9.99\times10^{-1}$ \\
$If   $      & $1.00\times10^0  $   & $1.00\times10^0   $ &$ 3.15\times10^{-4}$ & $9.99\times10^{-1}$ \\
$Mf  $       & $9.99\times10^{-1}$  & $9.99\times10^{-1}$ &$ 2.79\times10^{-4}$ &$ 9.99\times10^{-1}$ \\
$Tf      $   & $9.99\times10^{-1}$  & $9.99\times10^{-1}$ &$ 1.83\times10^{-4}$ & $9.99\times10^{-1}$ \\
$Tcf  $      & $1.00\times10^0$     & $1.00\times10^0   $ &$ 4.32\times10^{-4}$ & $9.99\times10^{-1}$ \\
\bottomrule
\end{tabular*}
\end{table}

\subsection{Synthetic data generation for training}
\label{SUBSEC:3_3}

The results of the MCMC allow the construction of the parameters PDF of phenomenological model capable of representing the uncertainty of the non-linear system. Thus, it is possible to build a set of non-linear responses by randomly selecting a set of parameters. In this work, a sample of 10000 distinct non-linear parameters was randomly selected. Subsequently, all resulting model were excited with the same LHS signal as in Fig. \ref{fig:06}, and the result was 10000 different dynamic responses.

\begin{figure}
     \centering
     \begin{subfigure}[b]{0.45\textwidth}
        \centering
        \includegraphics[width=1\columnwidth]{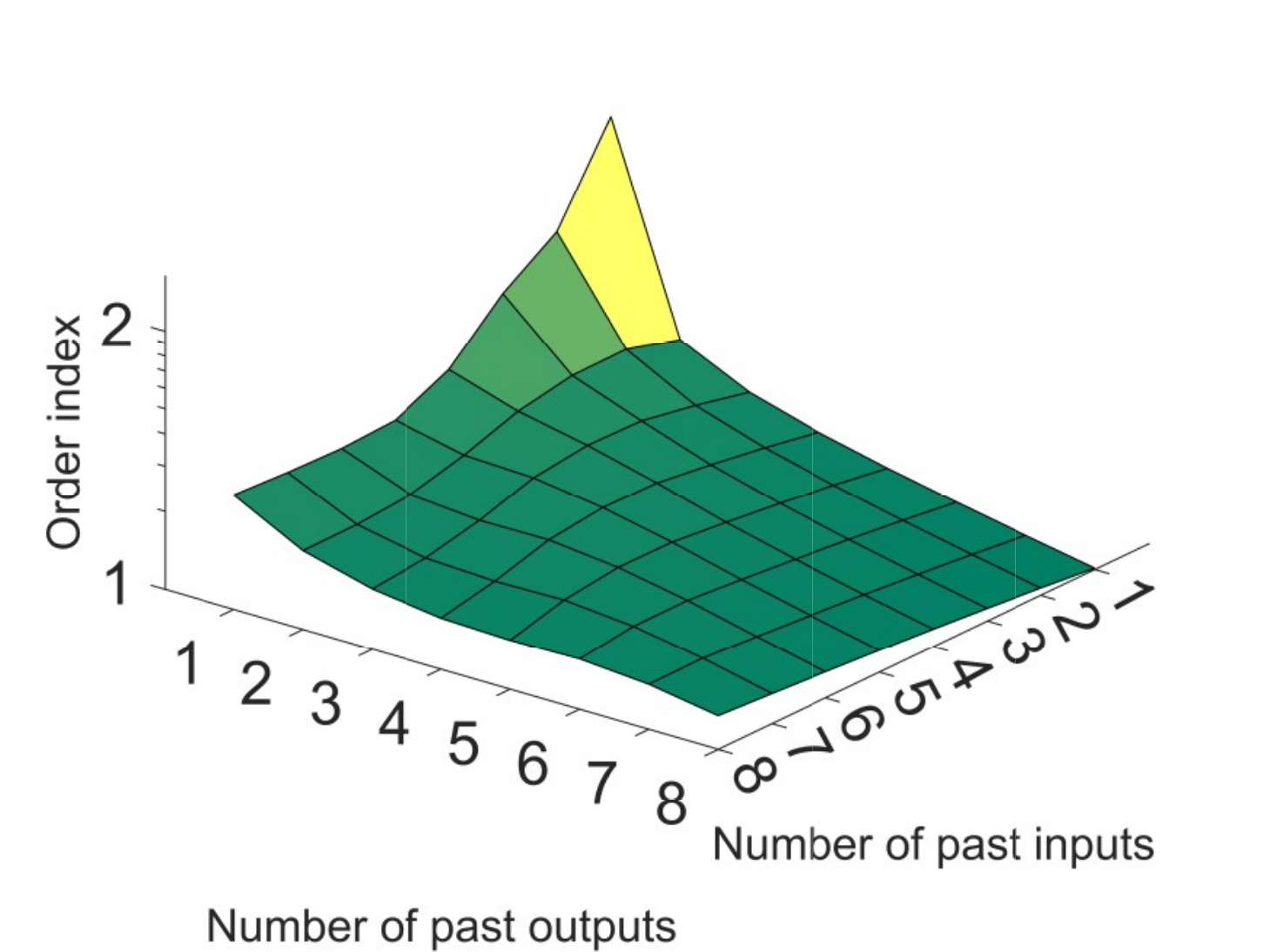}
        \caption{}
        \label{fig:09}
     \end{subfigure}
     \hfill
     \begin{subfigure}[b]{0.45\textwidth}
        \centering
        \includegraphics[width=1\columnwidth]{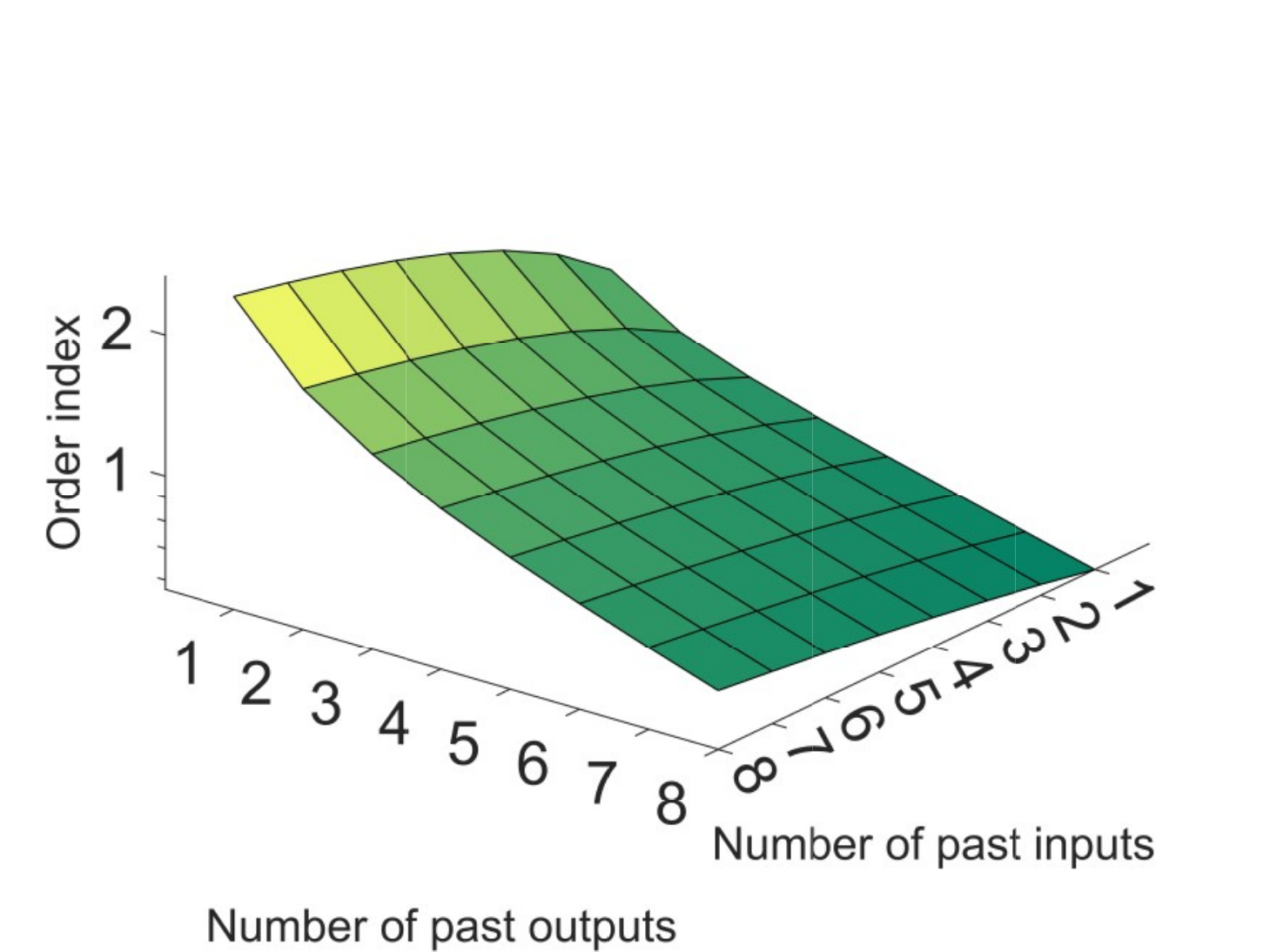}
        \caption{}
        \label{fig:10}
     \end{subfigure}
     \caption{a) Lipschitz surface for reactor temperature; b) Lipschitz surface for polymer viscosity.}
\end{figure}

With these dynamic trajectories, it was possible to establish the number of embedded dimensions of the NARX model. The Lipschitz method is presented in Sec. \ref{SUBSEC:2_3} and used to define the NARX parameters. Fig. \ref{fig:09} and \ref{fig:10} show these results. In Figures \ref{fig:09} and \ref{fig:10}, the decisive factor is the slope of the surface, because when there is a high variation between two delays, the increase is considered important. However, if the slope variation is low, it can be considered that this inclusion is not necessary. Then, it is possible to observe that a delay of four sampling instants for the inputs and one for the variables is reasonable for a good representation, as after these values, the slop starts to be constant.

\subsection{Monte Carlo training}
\label{SUBSEC:3_4}
The proposed methodology includes a step called Monte Carlo Training, composed of smaller steps. It starts with the identification of the Hyperparameters. It is followed by defining the data needed for training. Then, finally, the final Monte Carlo Training is done.

\subsubsection{Monte Carlo training} 
\label{SUBSUBSEC:3_4_1}

The network's optimal structure is found by a search in a hyperspace formed by the structural parameters that constitute the networks. Tab. \ref{tbl5} shows the initial configuration included in the Hyperband algorithm \cite{Li2016}. %In this way, the algorithm evaluated the network's architecture ranging from two to six dense layers; each layer could have between 30 and 160 neurons. Other structural parameters were found in Tab. \ref{tbl5}.

\begin{table}[width=.9\linewidth,cols=4,pos=h]
\caption{Hyperband hyperspace serach.}\label{tbl5}
\begin{tabular*}{\tblwidth}{@{} LL@{} }
\toprule
Parameter &	Search Space\\
\midrule
Type of layer &	Dense \\
Number of layers	& $2-6$\\
Output layer &	$1$\\\
Activation function	& $Relu$ or $Tanh$\\
Number of neurons per layer &	[30, 50, 70, 90, 100, 120, 130, 160]\\
Learning rate &	$0.0001$, $0.001$, and $0.1$\\
Metrics &	MAE – Mean Absolute Error\\
Loss &	MSE – Mean Square Error\\
\bottomrule
\end{tabular*}
\end{table}

The results obtained from the hyperparameters search are presented in Tab. \ref{tbl6}. It is possible to observe that a relatively simpler network was necessary to represent the temperature than viscosity. This simpler architecture implies a big difference in the computational cost. The viscosity network required 6.4 more times to be trained than the network for the temperature. On the other hand, networks have the same activation functions in the layers and the same learning rate. Tab. \ref{tbl6} also shows the MAE and MSE values resulting from the Hyperband search. The hyperband algorithm uses the training and validation datasets during the training. Then, the Test dataset is used after the training to test the final model parameters. It is noteworthy, however, that this methodology assumes that the network's architecture does not change to a variable. In this way, it is considered that it is only necessary to execute the hyperparameter search process once. With the network structure identified, it is trained for all trajectories obtained from each non-linear model. Hence, a set of networks are identified, as described in Sec. \ref{SUBSEC:2_5}.

\begin{table}[width=.9\linewidth,cols=4,pos=h]
\caption{Resulting networks hyperparameters.}\label{tbl6}
\begin{tabular*}{\tblwidth}{@{} LLL@{} }
\toprule
Hyperparameters & 	$T$	& $\eta$ \\
\midrule
Number of layers &	3 &	7 \\
Number of neurons in the dense layers	& $[100,90,1]$ &	$[150,90,150,90,150,90,1]$ \\
Activation function	& $[tanh, tanh]$ &	$[tanh, tanh, tanh, tanh, tanh, tanh]$ \\
Initial learning rate &	$1\times10^{-3}$ &	$1\times10^{-3}$ \\
The total number of trainable parameters &	$11081$ &$	71011$ \\
MSE Test &	$2.37\times10^{-5}$ &	$5.84\times10^{-4}$ \\ 
MAE Test &	$4.30\times10^{-3}$ &	$2.03\times10^{-2}$ \\
\bottomrule
\end{tabular*}
\end{table}

\subsubsection{Data size} 
\label{SUBSUBSEC:3_4_2}

\begin{figure}
    \centering
    \includegraphics[width=0.8\columnwidth]{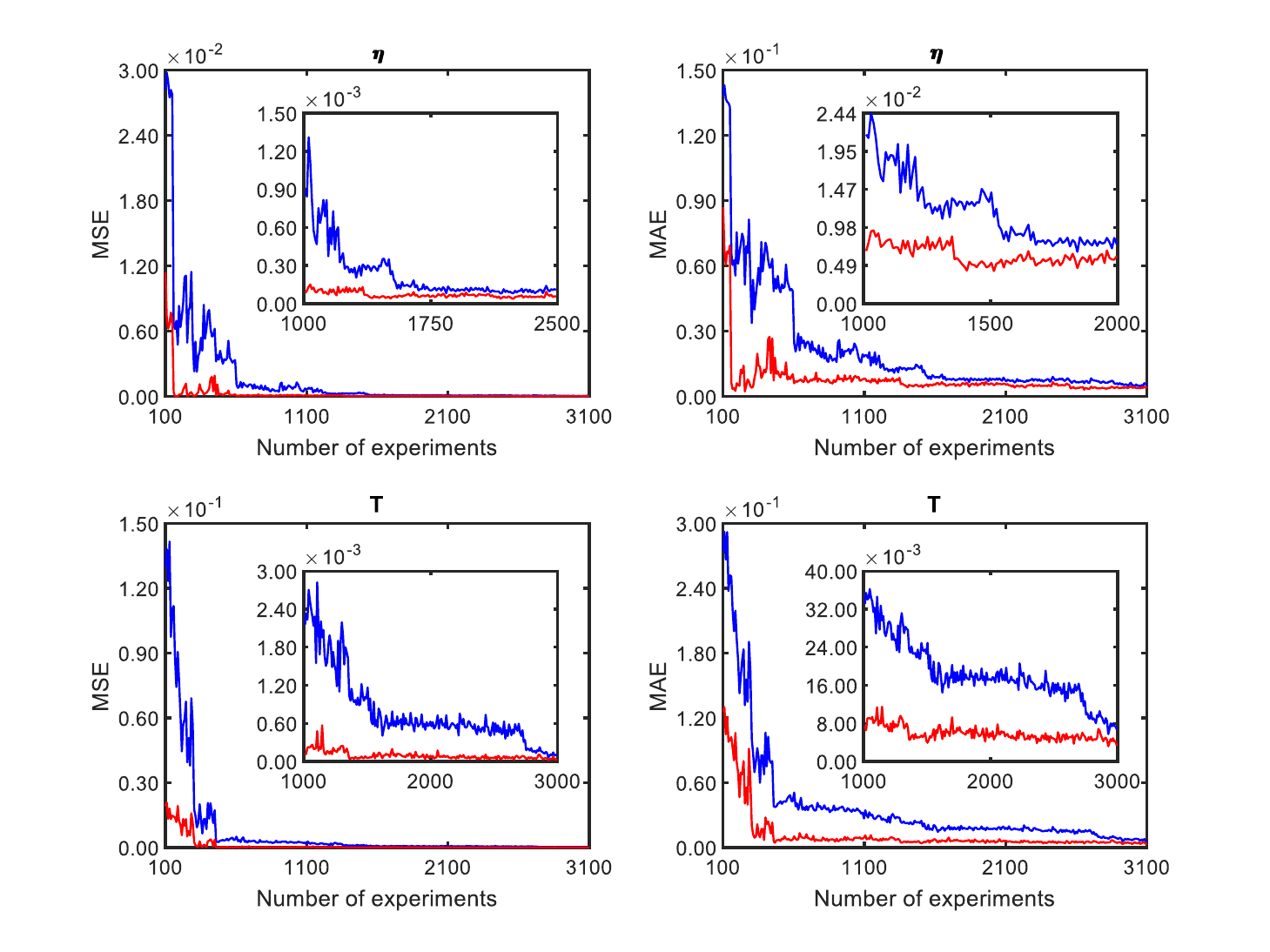}
    \caption{Training performance as a function of experiments.}
    \label{fig:11}
\end{figure}

An important aspect to be analyzed in neural network training is the guarantee of adequate training. In this sense, assessing whether the amount of information added in training the model is sufficient for the training algorithm to obtain a suitable model is necessary. Fig. \ref{fig:11} shows this analysis for the polymer viscosity and reactor temperature. This evaluation was based on a set of independent training in which each one was repeated twenty-five times. The first twenty-five training was carried out with 100 experiments. The average value MAE and the final MSE was calculated. For the next twenty-five, 100 experiments were added, and so on. In Fig. \ref{fig:11}, it is possible to observe that for viscosity, after about 1750 experiments, there is no significant change in the MSE; however, for MAE, this value is about 1500. On the other hand, when the reactor temperature is evaluated, this value is higher, and more than 2500 experiments are needed for convergence. Thus, to ensure convergence, 3100 were used for both networks.

\subsubsection{Training} 
\label{SUBSUBSEC:3_4_3}

The adaptive moment estimation algorithm (ADAM) proposed by \cite{Kingma2014} was used to train the chosen structure. The ADAM is an optimization method based on the descending gradient technique, making the ADAM algorithm efficient for problems involving extensive data and parameters. It also requires less memory than other training algorithms because the data is sliced into several packages and treated.

Building the networks involves an exhaustive training process called Monte Carlo Traning. The proposed methodology is based on the Monte Carlo method for PDF propagation. Thus, the assumed hypothesis is that the different trajectories generated through the non-linear model can represent the model's uncertainty. Therefore, training networks capable of representing these distinct trajectories implies obtaining a PDF of trained parameters of an AI model that also represent the uncertainty of the model.

\begin{figure}
    \centering
    \includegraphics[width=0.8\columnwidth]{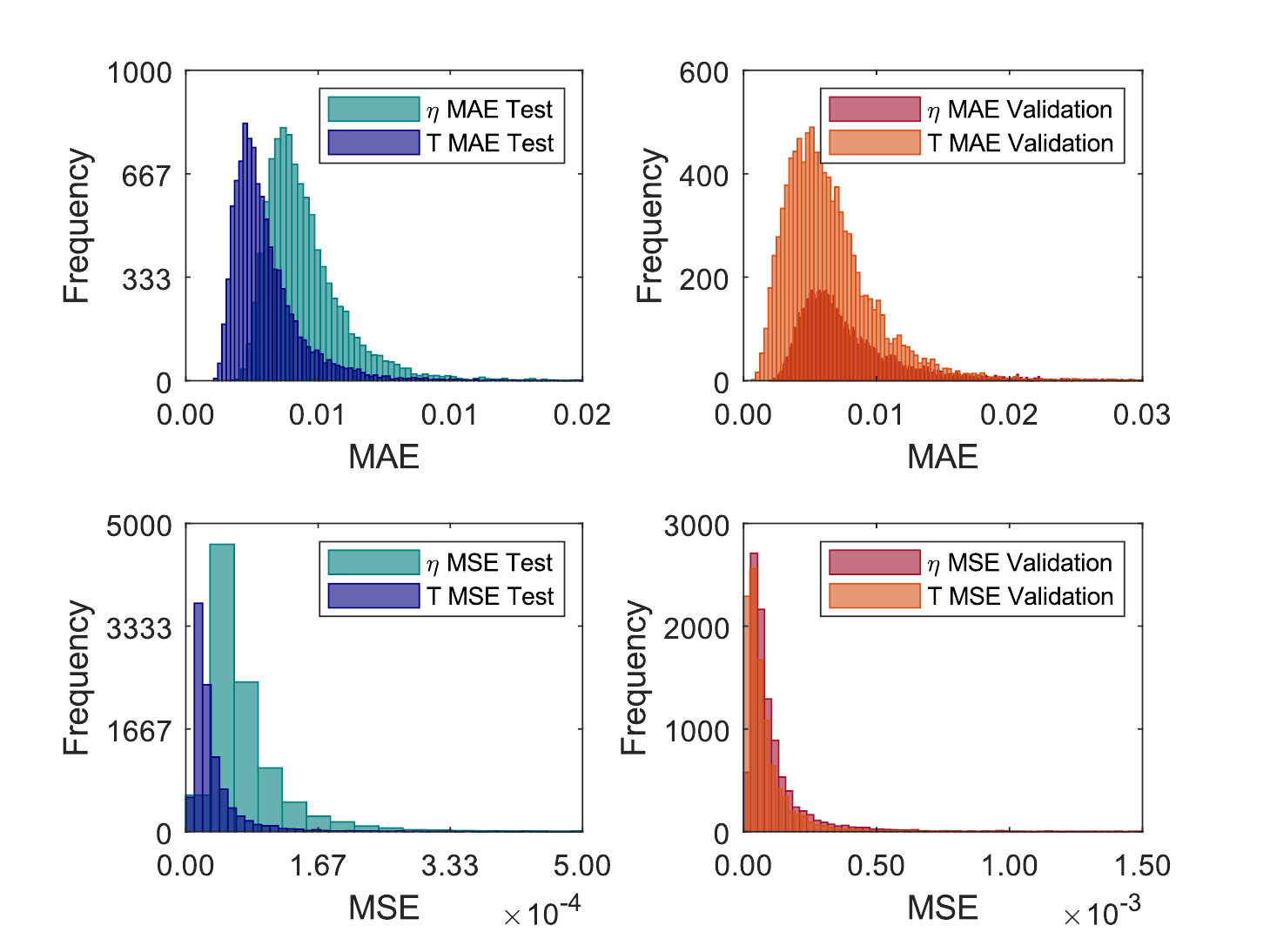}
    \caption{Training performance as a function of experiments.}
    \label{fig:12}
\end{figure}

Convergence analysis of training networks via MC Training can be performed by analyzing MAE and MSE values  like conventional training. However, given the number of trained networks, it is more convenient to evaluate in histogram format as in Fig. \ref{fig:12}. In a first analysis, Fig. \ref{fig:12} shows the histograms of the MAE and MSE indicators, both for the test and validation data. It is possible to observe that the histograms do not follow a Gaussian distribution. So, the mean may not be a good reference in statistical terms. Thus, Tab \ref{tbl7} shows the minimum, maximum, median, and the standard deviation of the MAE and MSE. Generally, it is possible to affirm that the networks converged sufficiently.

\begin{table}[width=.9\linewidth,cols=5,pos=h]
\caption{Resulting networks hyperparameters.}\label{tbl7}
\begin{tabular*}{\tblwidth}{@{} LLLLL@{} }
\toprule
& MAE Test &	MAE Valid &	MSE Test &	MSE Valid \\
\midrule
\multicolumn{5}{c}{\textbf{$T$}}  \\
\midrule
Min &	$1.49\times10^{-3}$ &	$7.91\times10^{-4}$&	$5.64\times10^{-6}$&	$2.73\times10^{-6}$ \\
Max	 & $2.26\times10^{-3}$&	$6.30\times10^{-2}$&	$1.05\times10^{-3}$&	$4.82\times10^{-3}$ \\
Median &	$3.57\times10^{-3}$&	$5.77\times10^{-3}$&	$2.37\times10^{-5}$&	$4.98\times10^{-5}$ \\
STD &	$1.82\times10^{-3}$&	$3.72\times10^{-3}$&	$4.37\times10^{-5}$&	$1.24\times10^{-4}$ \\
\midrule
\multicolumn{5}{c}{\textbf{$\eta$}}  \\
\midrule
Min &	$2.35\times10^{-3}$&	$2.06\times10^{-3}$&	$1.20\times10^{-5}$&	$1.06\times10^{-5}$ \\
Max &	$7.35\times10^{-2}$&	$2.01\times10^{-1}$&	$1.82\times10^{-2}$&	$5.27\times10^{-2}$ \\
Median &	$5.55\times10^{-3}$&	$6.79\times10^{-3}$&	$5.87\times10^{-5}$&	$7.22\times10^{-5}$ \\
STD &	$3.38\times10^{-3}$&	$6.23\times10^{-3}$&	$4.55\times10^{-4}$&	$9.25\times10^{-4}$ \\
\bottomrule
\end{tabular*}
\end{table}

\begin{figure}
    \centering
    \includegraphics[width=0.8\columnwidth]{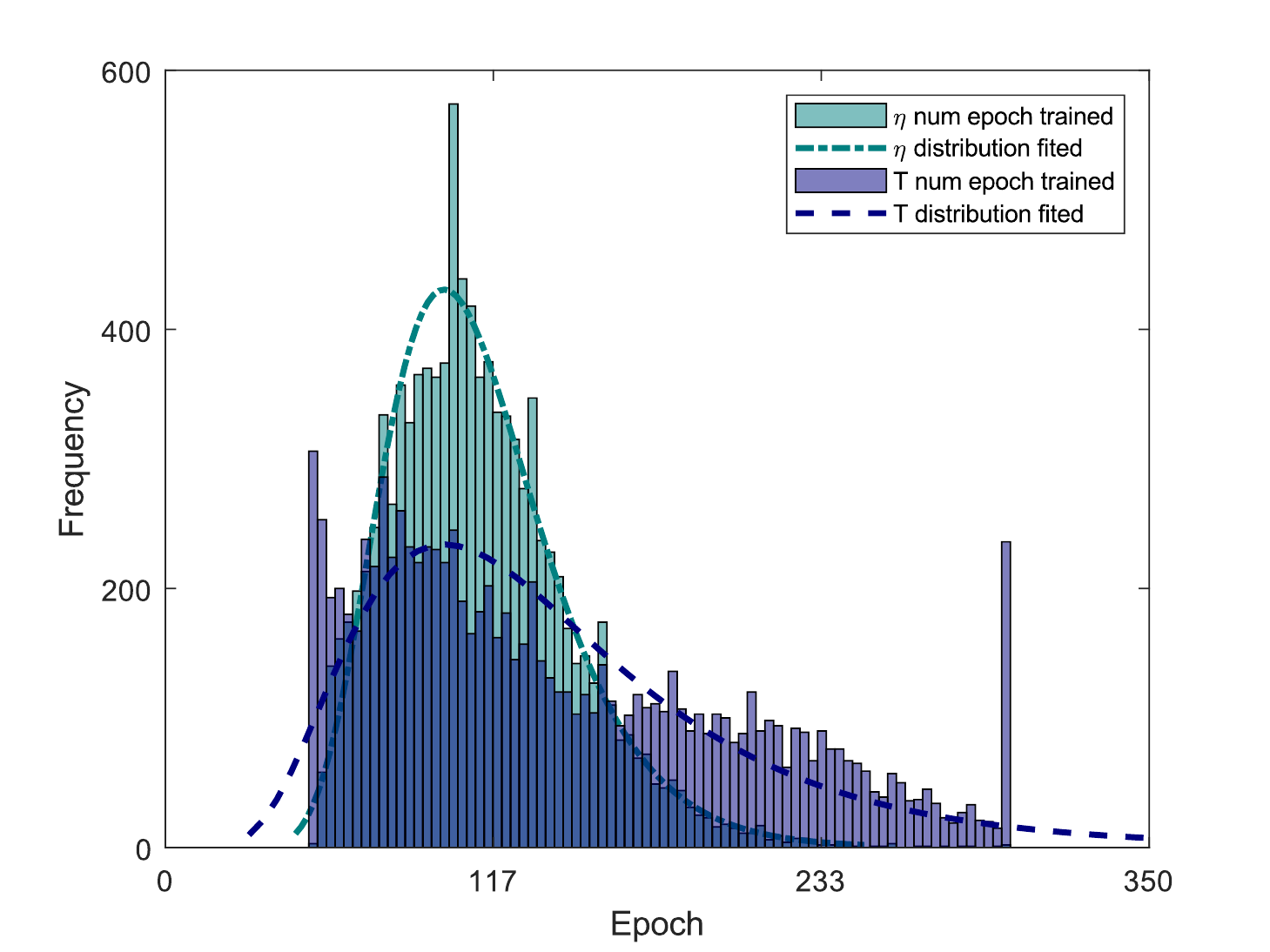}
    \caption{Trained epochs of Monte Carlo Training.}
    \label{fig:13}
\end{figure}

In the Monte Carlo Training process, an Early Stopping option in MC training was used to reduce the computational cost. Thus, training is aborted if there is no decrease in the LOSS value for 100 epochs. Fig. \ref{fig:13} shows the histograms of the number of epochs trained during MC training for the two modeled variables. In the graph of Fig. \ref{fig:13}, the minimum number of epochs was 50, and the maximum number was 300. Lognormal distributions should be the best representation for this data type with lower or upper bounds. Fig. \ref{fig:13} also shows that the data do not fit a lognormal distribution. Hence, no assumptions are made about the type of distribution of variables.

\subsection{Uncertainty propagation and Validation} 
\label{SUBSEC:3_5}

\begin{figure}[h!t]
    \centering
    \includegraphics[width=0.8\columnwidth]{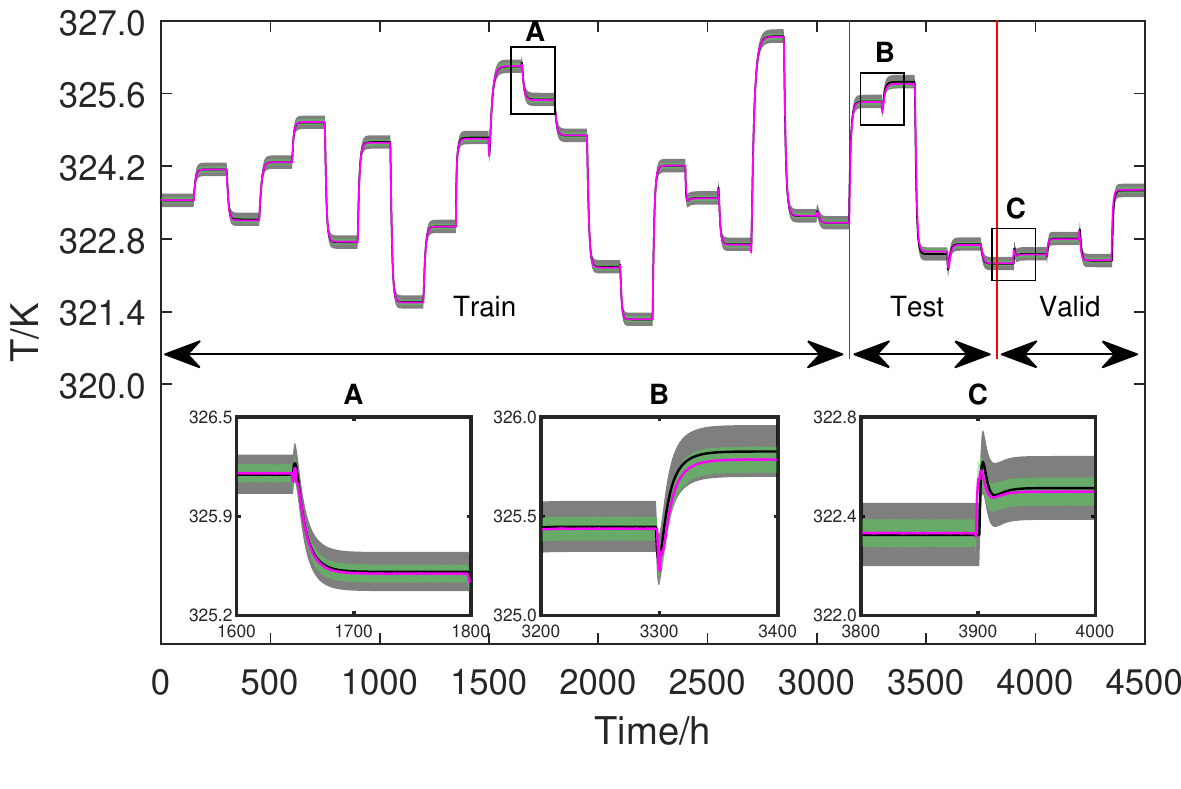}
    \caption{Training and validation data of $T$.}
    \label{fig:14}
\end{figure}

\begin{figure}[h!]
    \centering
    \includegraphics[width=0.8\columnwidth]{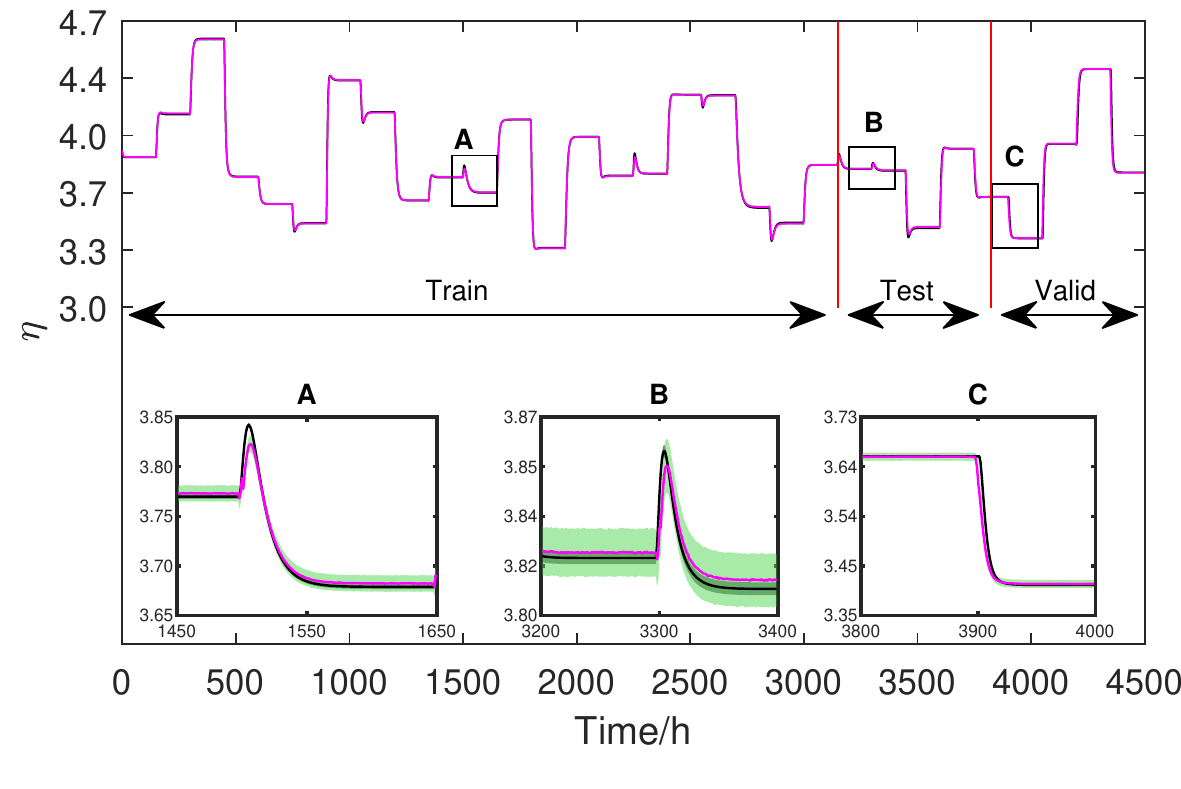}
    \caption{Training and validation data of $\eta$.}
    \label{fig:15}
\end{figure}

The last step of the proposed methodology is the propagation of uncertainty and methodology cross-validation. The proposal is based on constructing the uncertainty regions of the neural network's prediction and comparison with the uncertainty regions of the non-linear phenomenological model. With the prediction values, it is possible to calculate the variance of the networks using the same assumptions used to calculate the uncertainty of the phenomenological model Eq. \ref{EQ:13} to \ref{EQ:24}. Fig. \ref{fig:14} to \ref{fig:15} show the comparisons between the predictions for the two modeled variables, $T$, and $\eta$.

Fig. \ref{fig:14} to \ref{fig:15} present the entire output dataset used for training, testing, and validation. A zoom is given to verify the variables' behavior in each region. It is pointed out, however, that the training and validation data sets are used during the training of the networks. In this way, networks only unknown the test data set, which is used to certify their performance.

From a statistical point of view, when two measurements have overlapping coverage regions, it is impossible to differentiate, and they are considered equal. Therefore, the validation of the AI model against the non-linear phenomenological model is achieved when the dynamic range regions are superimposed. In this sense, the Graphs of Fig. \ref{fig:14} to \ref{fig:15} allow us to conclude that both models statistically produce the same dynamic response.

\section{Conclusion} 
\label{SEC:4}

This work presented a novel methodology for evaluating the uncertainty of Scientific Machine Learning Models. A comprehensive approach is proposed that considers several uncertainties associated with the SciML model structure, the data used, and the original data source. The proposed strategy was composed of five steps: Markov Chain Monte Carlo method to obtain the uncertainty of the non-linear model parameters; generate synthetic data; neural networks structure identification; Monte Carlo simulation training; methodology validation and uncertainty assessment of the trained mode.

The proposed method considers epistemic and aleatory uncertainties. These uncertainties are considered in the context of the data used to train the models and the model itself. Therefore, it is possible to provide an overall strategy for the uncertainty-aware models in the SciML field. A case study demonstrated the method's consistency. Hence, two Soft Sensors were identified to provide information about the temperature and viscosity of a polymerization reactor. The results indicated that the Soft Sensors predictions are statistically equal to the validation data in both dynamic and stationary regimes.

%%%%%%%%%%%%%%%%%

\printcredits

\bibliographystyle{cas-model2-names}

\bibliography{cas-refs}

\newpage
\appendix
%\begin{landscape}
\section{Supplementary data}
\label{APPEND}

\begin{figure}[h!]
    \centering
    \includegraphics[width=\columnwidth]{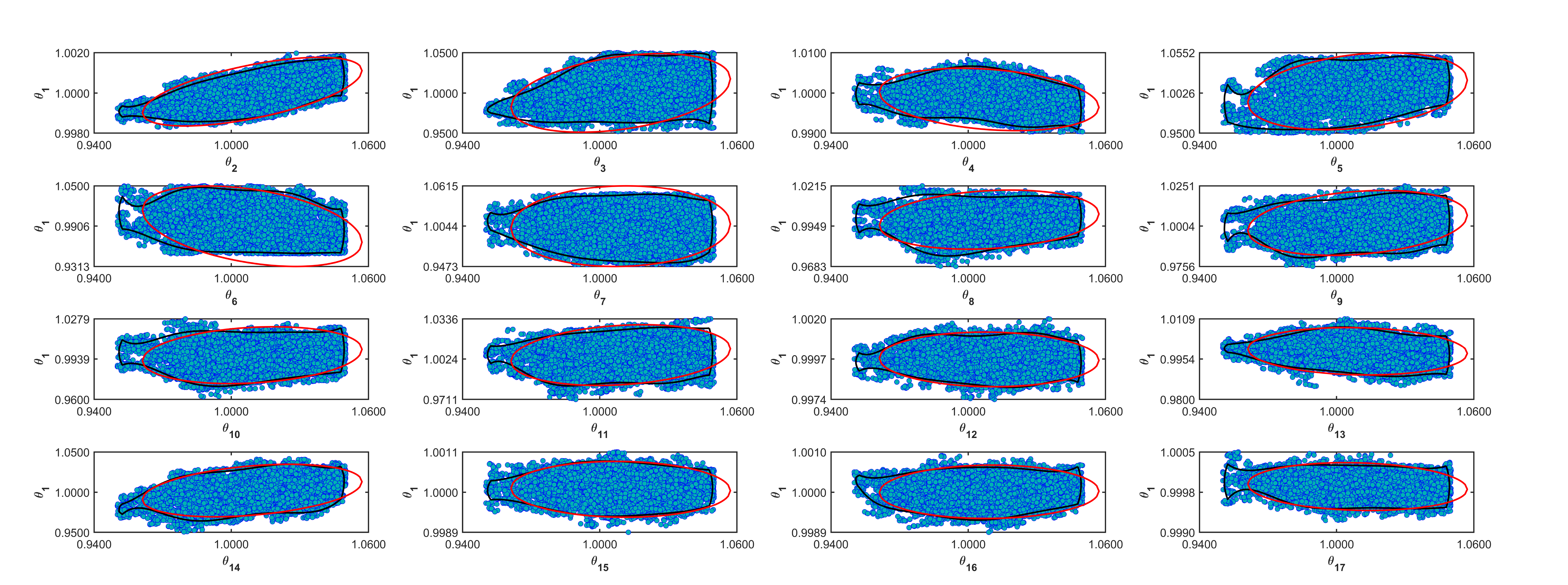}
    \caption{Coverage regions parameters 01. (\textcolor{red}{---}) Gaussian region; (\textcolor{black}{---}) \citet{Possolo2010} region.}
    \label{fig:Coverage_regions_1}
\end{figure}
\begin{figure}
    \centering
    \includegraphics[width=\columnwidth]{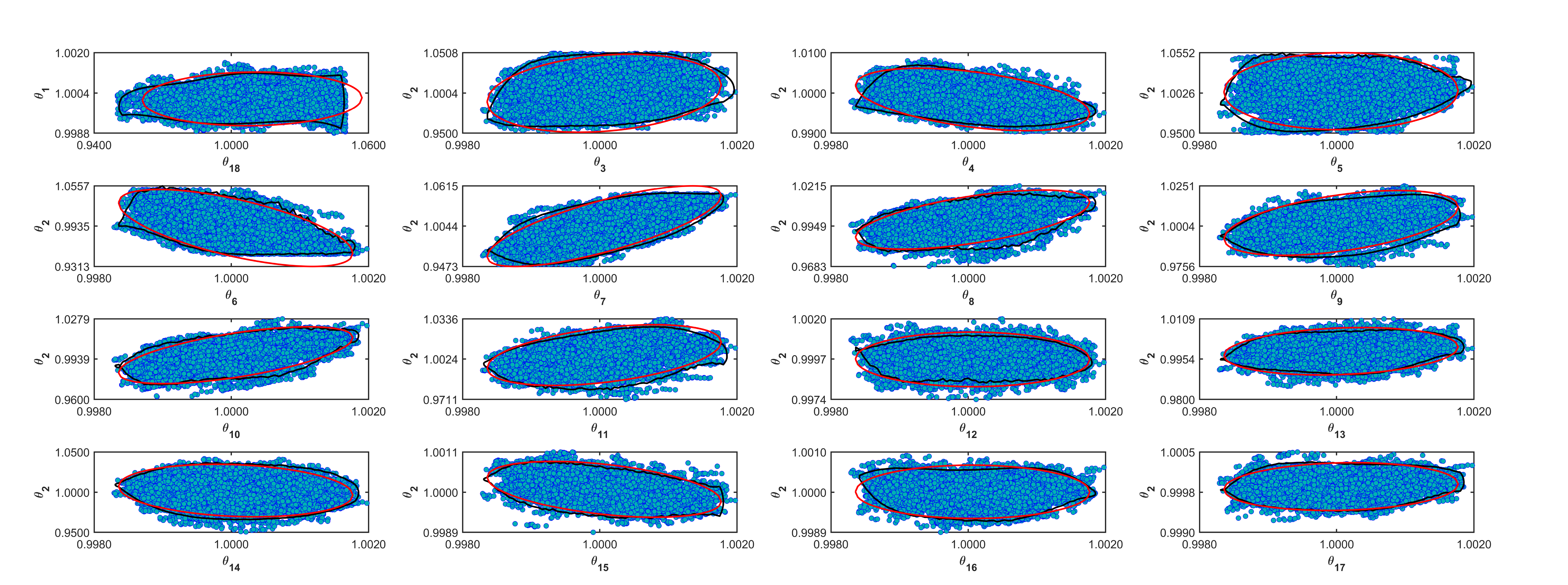}
    \caption{Coverage regions parameters 02. (\textcolor{red}{---}) Gaussian region; (\textcolor{black}{---}) \citet{Possolo2010} region.}
    \label{fig:Coverage_regions_2}
\end{figure}

\begin{figure}
    \centering
    \includegraphics[width=\columnwidth]{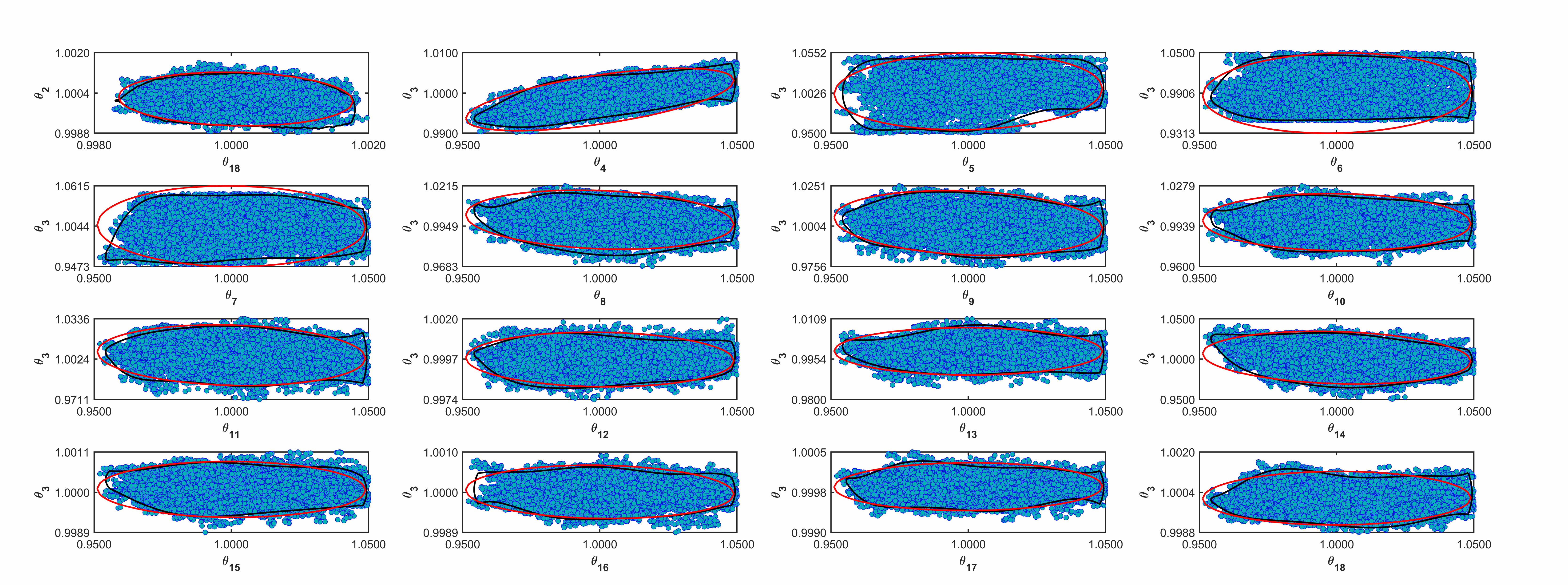}
    \caption{Coverage regions parameters 03. (\textcolor{red}{---}) Gaussian region; (\textcolor{black}{---}) \citet{Possolo2010} region.}
    \label{fig:Coverage_regions_3}
\end{figure}
\begin{figure}
    \centering
    \includegraphics[width=\columnwidth]{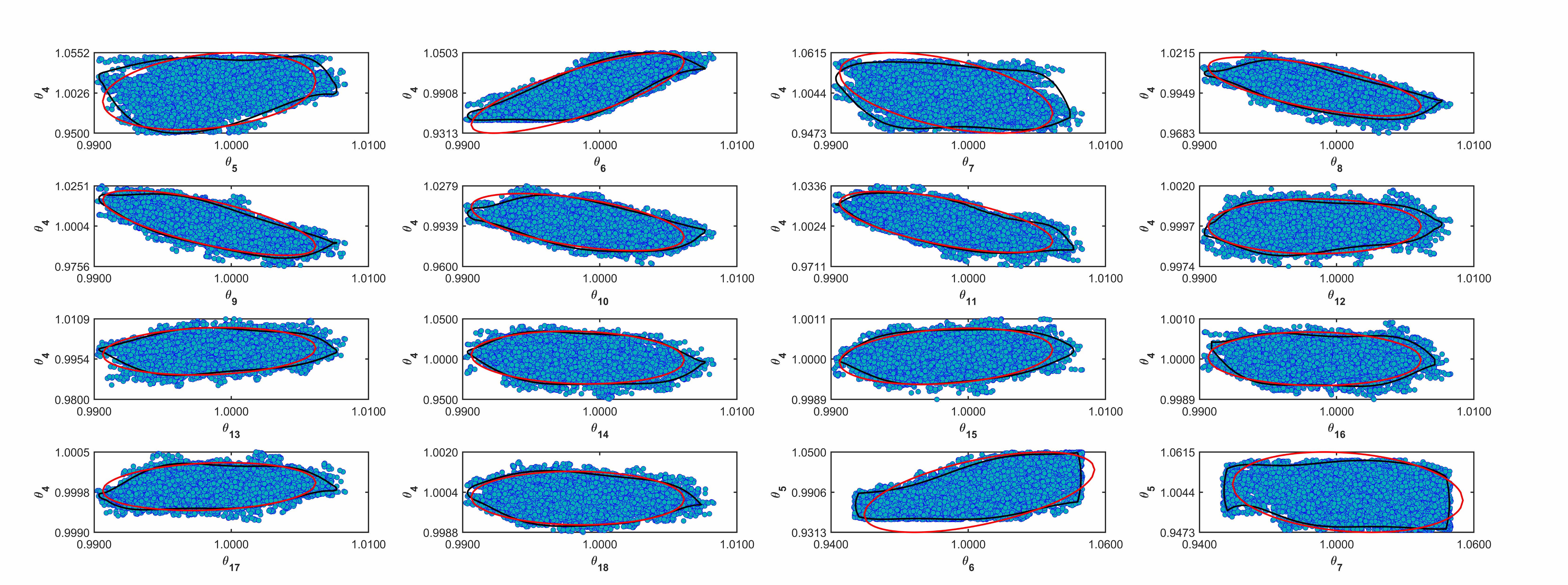}
    \caption{Coverage regions parameters 04. (\textcolor{red}{---}) Gaussian region; (\textcolor{black}{---}) \citet{Possolo2010} region.}
    \label{fig:Coverage_regions_4}
\end{figure}
\begin{figure}
    \centering
    \includegraphics[width=\columnwidth]{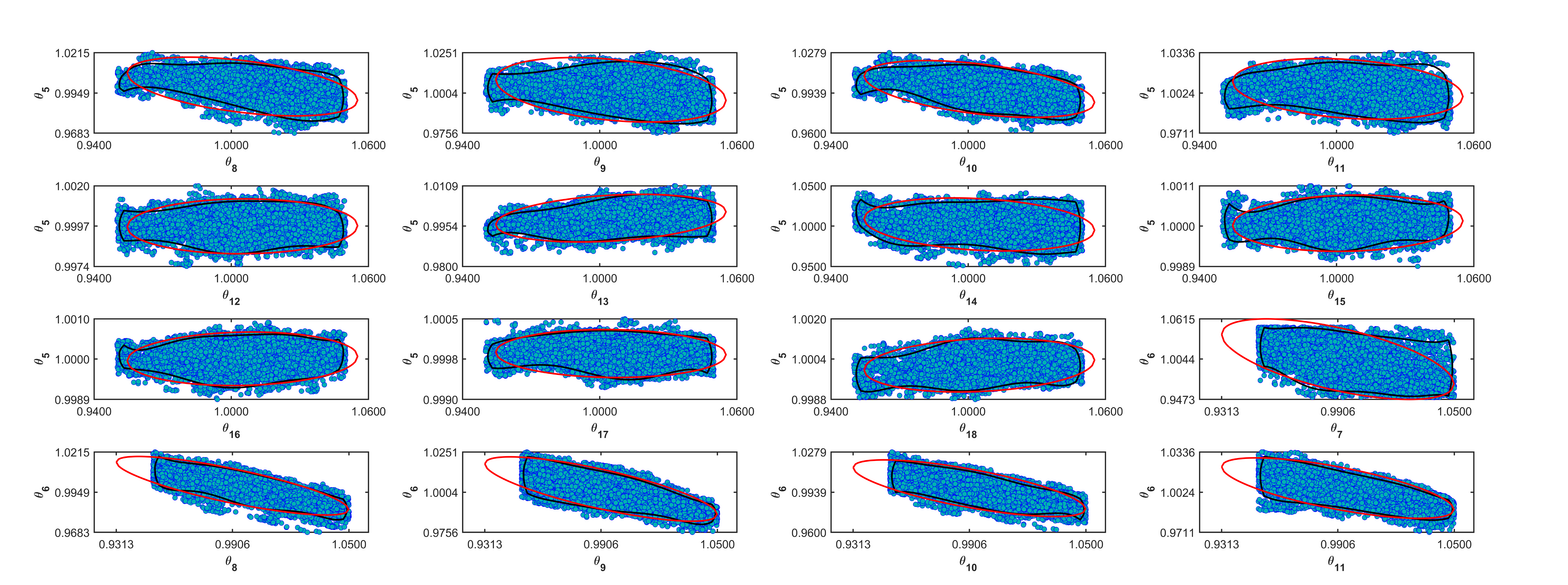}
    \caption{Coverage regions parameters 05. (\textcolor{red}{---}) Gaussian region; (\textcolor{black}{---}) \citet{Possolo2010} region.}
    \label{fig:Coverage_regions_5}
\end{figure}
\begin{figure}
    \centering
    \includegraphics[width=\columnwidth]{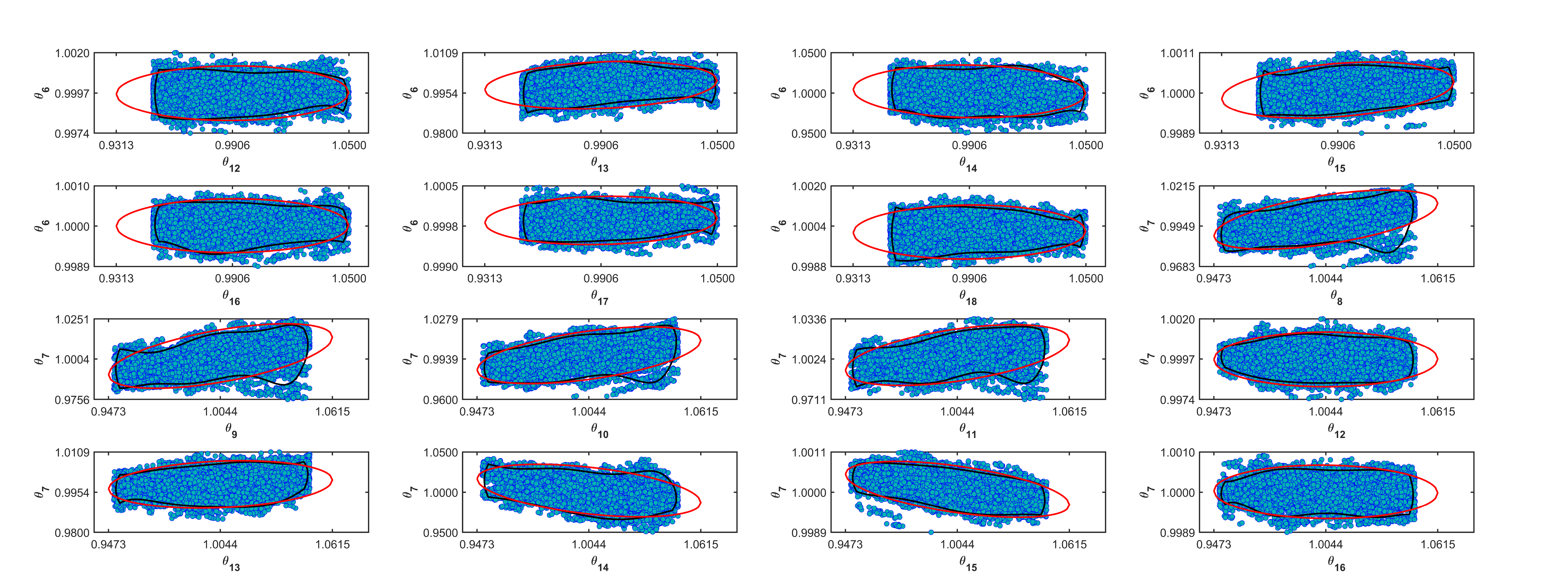}
    \caption{Coverage regions parameters 06. (\textcolor{red}{---}) Gaussian region; (\textcolor{black}{---}) \citet{Possolo2010} region.}
    \label{fig:Coverage_regions_6}
\end{figure}

\begin{figure}
    \centering
    \includegraphics[width=\columnwidth]{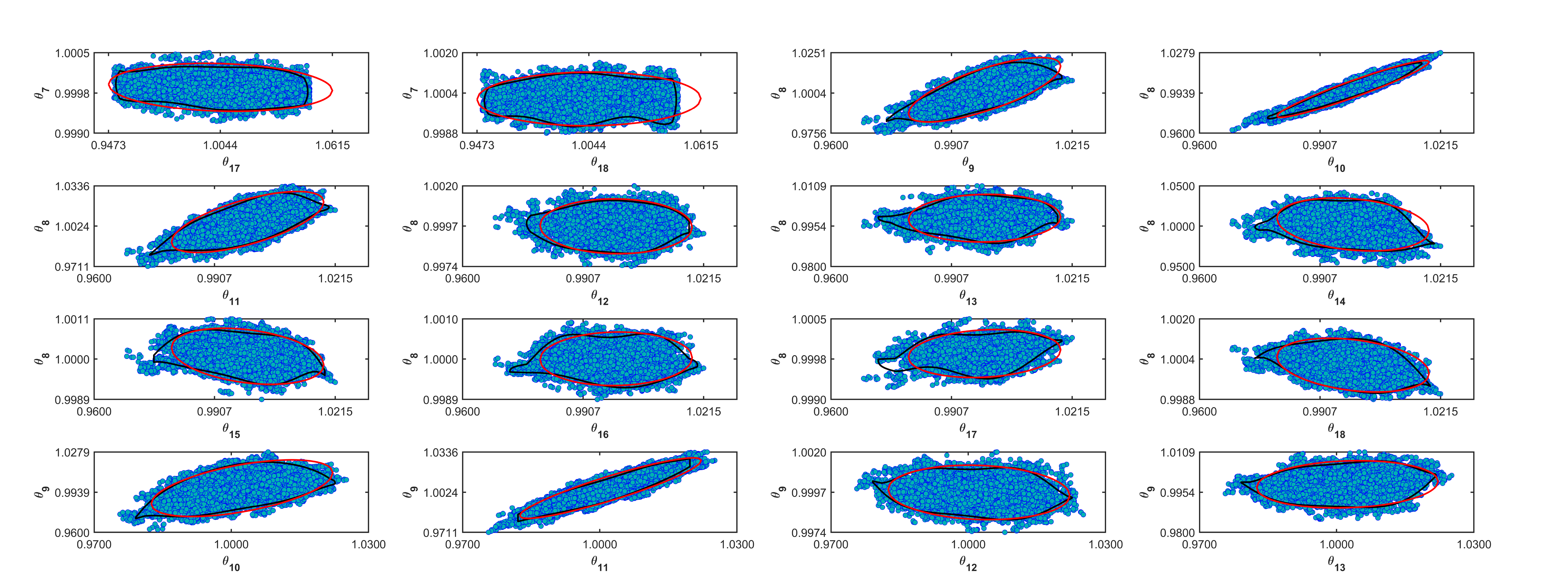}
    \caption{Coverage regions parameters 07. (\textcolor{red}{---}) Gaussian region; (\textcolor{black}{---}) \citet{Possolo2010} region.}
    \label{fig:Coverage_regions_7}
\end{figure}

\begin{figure}
    \centering
    \includegraphics[width=\columnwidth]{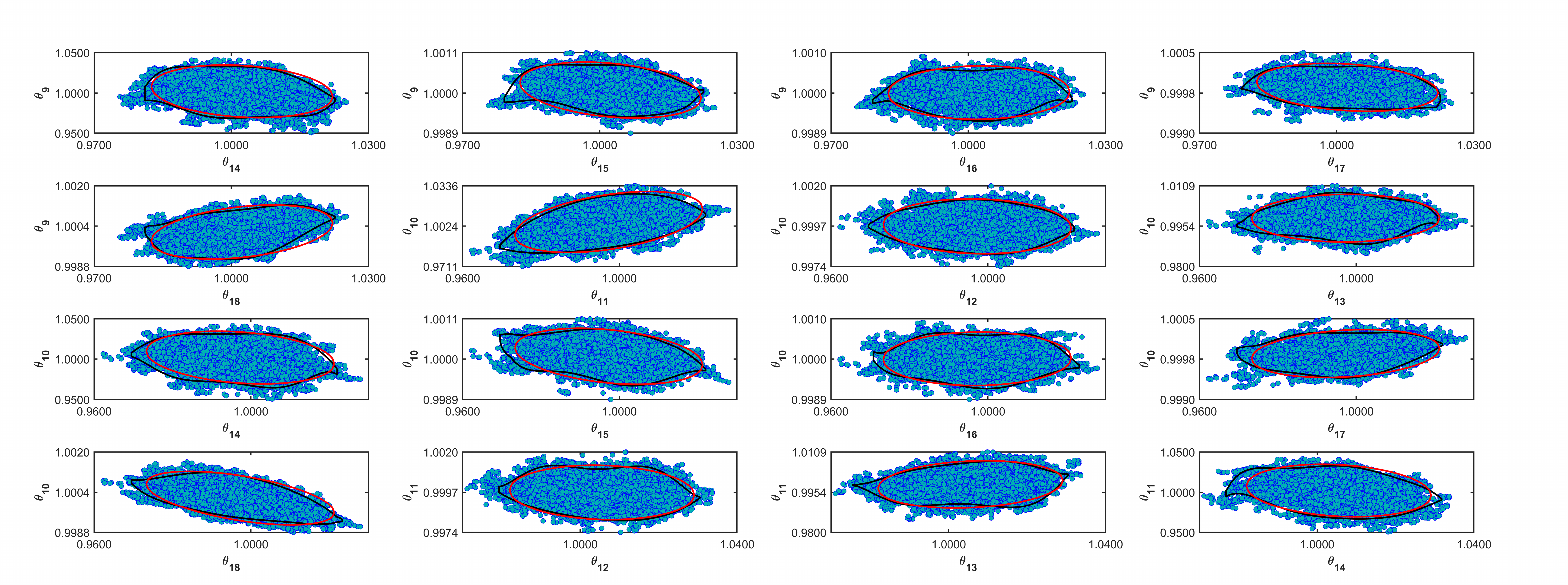}
    \caption{Coverage regions parameters 08. (\textcolor{red}{---}) Gaussian region; (\textcolor{black}{---}) \citet{Possolo2010} region.}
    \label{fig:Coverage_regions_8}
\end{figure}
\begin{comment}
\begin{figure}
    \centering
    \includegraphics[width=\columnwidth]{figs/Coverage_regions_9.pdf}
    \caption{Coverage regions parameters 09. (\textcolor{red}{---}) Gaussian region; (\textcolor{black}{---}) \citet{Possolo2010} region.}
    \label{fig:Coverage_regions_9}
\end{figure}

\begin{figure}
    \centering
    \includegraphics[width=\columnwidth]{figs/Coverage_regions_10.pdf}
    \caption{Coverage regions parameters 10. (\textcolor{red}{---}) Gaussian region; (\textcolor{black}{---}) \citet{Possolo2010} region.}
    \label{fig:Coverage_regions_10}
\end{figure}

\end{comment}
\begin{figure}
    \centering
    \includegraphics[width=\columnwidth]{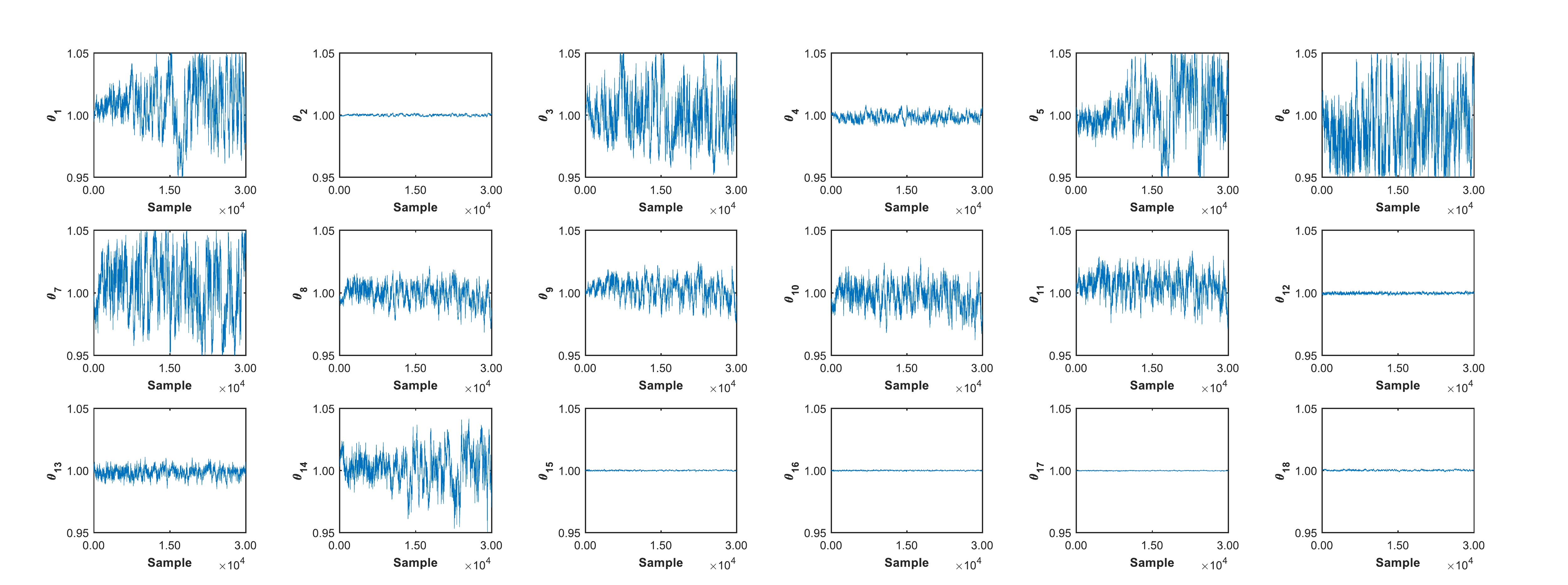}
    \caption{Parameters random walk.}
    \label{fig:Random_Walk_1}
\end{figure}
%\end{landscape}


%Author biography without author photo.
%Author biography. Author biography. Author biography.
%Author biography. Author biography. Author biography.

%\bio{figs/pic1}
%Author biography with author photo.
%Author biography. Author biography. Author biography.

%\bio{figs/pic1}
%Author biography with author photo.

\end{document}